\documentclass{ieeeaccess}
\usepackage{cite}
\usepackage{url}
\usepackage{amsmath,amssymb,amsfonts}
\usepackage{graphicx}
\usepackage{textcomp}
\usepackage{booktabs}
\usepackage{array}
\newcolumntype{R}[1]{>{\raggedright\arraybackslash}p{#1\textwidth}}
\newcommand{\zb}{\discretionary{}{}{}}

\usepackage{bm}
\makeatletter
\AtBeginDocument{\DeclareMathVersion{bold}
\SetSymbolFont{operators}{bold}{T1}{times}{b}{n}
\SetSymbolFont{NewLetters}{bold}{T1}{times}{b}{it}
\SetMathAlphabet{\mathrm}{bold}{T1}{times}{b}{n}
\SetMathAlphabet{\mathit}{bold}{T1}{times}{b}{it}
\SetMathAlphabet{\mathbf}{bold}{T1}{times}{b}{n}
\SetMathAlphabet{\mathtt}{bold}{OT1}{pcr}{b}{n}
\SetSymbolFont{symbols}{bold}{OMS}{cmsy}{b}{n}
\renewcommand\boldmath{\@nomath\boldmath\mathversion{bold}}}
\makeatother

\graphicspath{{../../figures/}}

\newcommand{\auccomp}{\ensuremath{\mathrm{AUC}_{\mathrm{comp}}}}
\newcommand{\ci}[2]{\ensuremath{[#1,\,#2]}}

\newcommand{\SM}{Supplementary Material}
\AddToHook{shipout/foreground}{\ifnum\ReadonlyShipoutCounter=1\relax
  \put(36,-31){\parbox[b]{400pt}{\normalfont\normalcolor\raggedright\hyphenpenalty=10000\scriptsize This work has been submitted to the IEEE for
  possible publication. Copyright may be transferred without notice, after which this version may no longer be accessible.}}\fi}
\begin{document}
\history{Date of publication xxxx 00, 0000, date of current version xxxx 00, 0000.}
\doi{10.1109/ACCESS.2026.0000000}

\title{A Controlled, Internally Audited Negative Result: Training Recipe, H.264
Compression Levels and a Frequency-Stream Deepfake Detector}

\author{\uppercase{Abbas Aliyev}\authorrefmark{1}
and \uppercase{Samir Rustamov}\authorrefmark{1,2}}

\address[1]{School of Information Technologies and Engineering, ADA University,
Baku AZ1008, Azerbaijan (e-mail: ashaliyev@ada.edu.az; srustamov@ada.edu.az)}
\address[2]{MegaSec AI Company, Baku, Azerbaijan}

\tfootnote{This work was developed within the Center of Data Analytics
Research (CeDAR) and the MegaSec AI Laboratory at ADA University.}

\markboth
{Aliyev, Rustamov: A Controlled, Internally Audited Negative Result}
{Aliyev, Rustamov: A Controlled, Internally Audited Negative Result}

\corresp{Corresponding author: Abbas Aliyev (e-mail: ashaliyev@ada.edu.az).}

\begin{abstract}
We test CAFRL, which adds block-DCT and
FFT-phase streams with compression-conditioned band attention, and an
adversarial compression-invariance loss, to an EfficientNet-B0 face backbone
for deepfake detection. Decision rules were frozen before the comparisons they
govern were run; controls were matched on data, augmentation and, after an
internal audit, training recipe; contrasts are paired on test videos. (1)~Configuration
result (registered STOP-gate, one seed): CAFRL as specified scored 3.66 AUC
points below a backbone-matched EfficientNet-B0 at CRF 40 --- as configured,
not an architectural effect. (2)~Recipe sensitivity: the control's
recipe recovered 3.96 points for CAFRL's
own components (post-audit pre-specified); of changes reverted singly (one seed,
descriptive), only the backbone learning rate's interval excluded zero, which
replicated on two further seeds (exploratory; a sensitivity, not a causal
decomposition). (3)~Neutral result: at that baseline-derived recipe, the frequency-stream
variants, not separately tuned, showed no detectable
difference at the resolution of this experiment from the spatial control, under the evaluated configurations (post-audit
pre-specified); a standalone frequency head scored 85.27 points at CRF 40
(exploratory). (4)~One-configuration neutral result: the adversarial loss, restored at
one strength, made no detectable difference (registered; 0.87
points, Holm-adjusted $p = 0.084$). (5)~Registered three-seed results:
real H.264 outperforms JPEG at matched mean PSNR, by 7.48 points at CRF 40 (averaged over CRF-40 exposure),
and CRF-40-level data added 1.88 points averaged over codec. These findings, and the advice to train on real
compression levels, cover FaceForensics++-family, GAN-era, single-pass H.264
data; diffusion content, transcoding chains and unknown codecs are untested.
\end{abstract}


\titlepgskip=-21pt

\maketitle

\section{Introduction}\label{sec:intro}

\PARstart{D}{etecting} manipulated face video in the wild means detecting it
\emph{after} social platforms have re-encoded it. H.264 compression at high
constant-rate-factor (CRF) settings destroys exactly the high-frequency
residue that many detectors rely on, and detectors trained on lightly
compressed (c23) data degrade on heavily compressed (c40) versions of the same
content --- a collapse documented by the standard
benchmark protocols \cite{roessler2019faceforensics,yan2023deepfakebench}. Under
those protocols our reproductions of three published detectors lose 12.8
(Xception), 13.2 (F3-Net) and 21.9 (FreqNet) points of video-level area under the receiver operating
characteristic curve (AUC) from c23 to c40
(\SM~Table~\ref{tab:appbench}). A large body of work responds with a common
prescription: extract frequency-domain features (DCT spectra, FFT phase, band
statistics), and make the representation \emph{invariant} to compression,
typically with adversarial objectives borrowed from domain adaptation
\cite{ganin2015unsupervised,cao2021metric,ma2024sensitivity,li2025plada}. The
intuition is compelling --- generative pipelines leave spectral fingerprints
\cite{frank2020leveraging,durall2020watch}, and invariance should preserve
what survives re-encoding.

This paper is a methods study of one instance of that prescription: how it
was tested under strict controls, how the negative result it produced was
internally audited, and what survived. \textbf{CAFRL}
(Compression-Aware Frequency Representation Learning) combines a block-DCT
stream and an FFT-phase stream with an ImageNet-pretrained spatial backbone; a
learned compression-level estimator conditions an adaptive frequency-band
attention gate (in the style of feature-wise linear modulation, FiLM
\cite{perez2018film}); and a gradient-reversal
discriminator enforces compression invariance on the frequency representation.
The design was assembled from a review of the compression-robust detection
literature (Section~\ref{sec:rationale}), and every ingredient in it is
individually prescribed there: DCT-band mining \cite{qian2020f3net}; the
phase spectrum, which carries up-sampling traces \cite{liu2021spsl}; FiLM-style
conditioning \cite{perez2018film}; cross-modal attention fusion
\cite{zhao2023cmam,luo2021generalizing}; and gradient-reversal compression
invariance \cite{ganin2015unsupervised,cao2021metric,ma2024sensitivity,li2025plada}.
Their combination, and the training recipe that carries it --- loss weights,
curriculum, per-group learning rates --- is a plausible assembly we
constructed from common ingredients, not a configuration any prior work
prescribes. The frequency representation under test is computed after
decoding, on the geometrically warped face crop: an $8{\times}8$ block DCT of
its luma and the phase of its FFT, not the codec's own coefficient domain.
What this study tests is therefore this
representation and its fusion, not frequency-based detection in general
(Limitation~(ii)).

\textbf{Research question.} We do not ask whether frequency representations
or compression invariance are useful in general. We ask a narrower question:
do the specific CAFRL frequency pathways --- the block-DCT and FFT-phase
streams with their estimator-conditioned band gate, fused with an
EfficientNet-B0 trunk --- provide incremental discrimination over
appropriately controlled spatial architectures when training includes
multiple H.264 compression levels? Two secondary questions are tested in a
registered revision campaign: whether CAFRL's adversarial
compression-invariance loss contributes once the training recipe is matched,
and on which part of the multi-level training data the result at c40
depends.

\textbf{How it was answered.} Each trained model configuration is an
\emph{arm}. Before any ablation ran, we froze a paired-bootstrap win-rule and
a STOP-gate: the full model had to beat backbone- and augmentation-matched
baselines before an ablation campaign would be spent. The gate fired against
CAFRL as specified (Section~\ref{sec:falsification}). We then audited our own
negative before believing it (Section~\ref{sec:audit}): the audit found eight
defects, D1--D8, four of them biased against the frequency hypothesis. Three
post-audit pre-specified re-tests repaired them, and a fourth, evaluation-only,
retired D5; two further seeds of each
re-tested frequency arm left the comparison's reading unchanged (Section~\ref{sec:retests}). The revision campaign, registered under a second
tag, then restored the adversarial loss at the matched recipe (arm E1, tests
T1 and T2), varied the training data (tests T3--T5), and reverted five of
the recipe changes one at a time (Section~\ref{sec:campaign}). Evidence tiers --- by timing \emph{registered},
\emph{post-audit pre-specified} or \emph{post hoc} --- are kept apart
throughout (Section~\ref{sec:tiers}), and the contrasts the claims rest on
are paired on test videos, with a primary reading that resamples seeds as well as test
videos (added post hoc for the original contrasts; registered for the
revision campaign).

\textbf{What it found, in brief.} The registered comparison went against CAFRL
as specified, in one seed-42 run per arm --- a configuration result, as
configured, not an architectural effect. Changing to the control's training recipe recovered that
deficit for CAFRL's own components, though it does not show that every component is irrelevant;
of the recipe changes reverted singly, only the backbone learning rate's
interval excluded zero, and this replicated on two further seeds (exploratory;
a sensitivity, not a causal decomposition). At the
matched recipe, these implementations of the frequency streams, not
separately tuned and run on a baseline-derived recipe, made no detectable
difference under the evaluated configurations, and neither did the adversarial loss when it was restored. Of the
training-data changes tested, real H.264 outperforms JPEG at matched mean PSNR
(peak signal-to-noise ratio; JPEG versions of the same frames), on
FaceForensics++-family, single-pass H.264 data. Section~\ref{sec:conclusionfinal} states these as
five separate findings, each with the evidence it rests on and its limit.

\subsection{Contributions}
\begin{itemize}
\item \textbf{C1 --- a protocol for testing, and auditing, a negative
architecture result:} a registered STOP-gate and paired-$\Delta$ win-rules,
backbone- and augmentation-matched controls, a trunk control trained under the
comparator's recipe (TrunkCtrl), an \emph{internal audit of the negative} with
post-audit pre-specified re-tests, evidence tiers kept apart, a primary
reading that resamples seeds and then source-identity clusters (added post hoc
for the original contrasts; registered for the revision campaign), and a
registered campaign of five confirmatory tests. Every FaceForensics++ (FF++) contrast of the
original analyses is released with the per-video scores that let it be
re-adjudicated; the out-of-distribution cells are released as summary
statistics only, where they were saved (Sections~\ref{sec:protocol}, \ref{sec:audit},
\ref{sec:discussion}; \SM~\S\ref{app:repro}).
\item \textbf{C2 --- a controlled, internally audited result for one frequency design},
in two parts, because no one experiment establishes both. \emph{(a)} Against a
backbone-matched control, at matched augmentation and training recipe, and at
two spatial-feature widths, block-DCT and FFT-phase streams under cross-modal attention
(CMAM) or $\gamma$-residual fusion showed no detectable difference from a plain spatial
backbone on multi-level, single-pass H.264 FF++ data, at the resolution of
this experiment and under the evaluated configurations: these implementations
were not separately tuned and ran on a baseline-derived recipe
(Limitation~(iii)). A standalone head shows
that the frequency sub-network is discriminative; whether what it carries is
already in the trunk, or the fusion does not exploit it, is not resolved and
is the main open question (Section~\ref{sec:redundancy}). \emph{(b)} The adversarial
compression-invariance loss added nothing detectable at CAFRL's shipped recipe
and degraded the estimator it conditions (D6). Restored at the matched recipe
(E1, one strength, three seeds), it made no detectable difference to detection
at c40 or averaged over the four heaviest levels (T1, T2), while the estimator stayed accurate at the
selected checkpoints; smaller effects in either direction are not excluded
(Section~\ref{sec:adversary}). Both parts hold at one operating point: a
4--21\,M CNN on FaceForensics++, ${\approx}0.96$ intra-dataset and
${\approx}0.70$ cross-dataset on Celeb-DF~v2, with DFDC and WildDeepfake only
in exploratory analyses (Sections~\ref{sec:falsification}--\ref{sec:retests}).
\item \textbf{C3 --- a training-data result, within its limits:} in registered
three-seed tests paired on test videos, real H.264 outperforms JPEG at matched
mean PSNR: training the plain EfficientNet-B0 control on real
H.264 levels gave a higher mean single-model c40 AUC than training on JPEG
versions of the same uncompressed frames matched on mean crop PSNR, by 7.48\,pt averaged over c40 exposure (T4),
and including the c40-level data in training raised it by 1.88\,pt averaged
over codec (T5; raw weights). Trained without c40, the control kept
$R = 0.88$ \ci{0.75}{1.03} of its c40 gain, not detectably below 1, with a
smaller loss not excluded (T3). This neutral test, read apart from T5, establishes
neither that c40-level exposure is dispensable nor that it is indispensable; the
flat degradation curve is measured within the trained single-pass H.264 levels and on
FaceForensics++ test data, and it does not carry to the near-domain corpus at
the same CRF levels (Sections~\ref{sec:data} and~\ref{sec:datatests}). The
submitted single-run comparison with JPEG augmentation is post hoc and
unpaired, and this claim does not rest on it.
\item \textbf{C4 --- harmonized benchmarks:} video-level compression sweeps
for Xception, F3-Net (our reimplementation of its frequency-aware decomposition
branch, FAD, see
\SM~\S\ref{app:baselines}), FreqNet (to our knowledge its first video-level
c40 report, and the one arm trained from scratch rather than
ImageNet-initialized) and EfficientNet-B0 on one harness, tabulated at the
compressed levels in \SM~Table~\ref{tab:appbench} with the full seven-level matrix
and the per-video predictions released
(Section~\ref{sec:falsification}, \SM~\S\ref{app:baselines}).
\end{itemize}

\subsection{What this study covers}
FaceForensics++-family evidence: FF++
training and intra-dataset testing, one held-out manipulation (FaceShifter), a
\emph{near-domain} check on DeepFakeDetection (DFD) --- which shares FF++'s capture and processing
lineage rather than being an independent corpus --- and a cross-dataset check
on Celeb-DF~v2, with DFDC \cite{dolhansky2020dfdc} and WildDeepfake
\cite{zi2020wilddeepfake} evaluated only in an exploratory analysis added in
revision, on a fixed DFDC subset and the WildDeepfake test split
(\SM~\S\ref{app:exploratory}). Every manipulation is GAN- or
graphics-era (2019--2020) and compression is single-pass H.264 CRF
re-encoding, apart from an exploratory test-time check with H.265 and AV1
(\SM~\S\ref{app:exploratory}). Section~\ref{sec:discussion}'s Limitations state which conclusion
each of these bounds and mark the untested cells.

\section{Related work}\label{sec:related}

\subsection{Frequency-domain deepfake detection}
F3-Net \cite{qian2020f3net}
mines frequency-aware clues with DCT-based decomposition; SPSL
\cite{liu2021spsl} shows the phase spectrum is more sensitive to up-sampling
than the amplitude spectrum; FreqNet \cite{tan2024freqnet} learns source-agnostic
high-frequency representations. Spectral fingerprints of generative
up-sampling are documented by \cite{frank2020leveraging,durall2020watch}.
A parallel lineage uses noise residuals rather than transform coefficients:
Luo et al.\ \cite{luo2021generalizing} extract spatial rich model (SRM) high-frequency features
from intermediate feature maps and fuse them with an RGB Xception trunk
through dual cross-modality attention, reporting cross-manipulation gains
\emph{on top of} a strong spatial trunk --- the closest published
counter-evidence to the no-detectable-marginal-value reading we report below,
and a frequency family our instantiation does not test. Its gains are large (up to
${+}0.29$ AUC cross-manipulation, ${+}0.20$ on Celeb-DF~v2) and are measured
at one fixed quality, FF++ c23, which is exactly the regime
\SM~Table~\ref{tab:controls} identifies as the one this paper does not test in. These works motivate frequency
features under compression but largely evaluate at light compression; our
study asks whether one such frequency path adds incremental discrimination
when the training data already spans several H.264 compression levels
(Section~\ref{sec:intro}).

\subsection{Compression-robust detection}
Cao et al.\ \cite{cao2021metric}
combine metric and adversarial learning for compression-insensitive forgery
features; Sensitivity Decouple Learning \cite{ma2024sensitivity} separates
compression-insensitive from compression-sensitive components --- an image
\emph{restoration} method rather than a detector, whose decoupling mechanism
this line of work borrows; PLADA
\cite{li2025plada} pairs a gradient-reversal compression-invariance objective
with a compression-detection head in the JPEG image domain --- the closest
prior to CAFRL's adversarial component. DANet-ATP \cite{yuan2025danet} applies
adversarial compression-invariance to audio deepfakes. CAFRL differs in
operating on H.264 \emph{video} with hybrid DCT+phase streams and
\emph{ordinal compression-level} conditioning rather than invariance alone; we
do not claim priority for adversarial compression-invariance itself. The
intuitions under test are current rather than historical: the works
instantiating them here are published in 2024--2025 venues
\cite{tan2024freqnet,ma2024sensitivity,li2025plada,yuan2025danet}.
\SM~Table~\ref{tab:protocols} classifies the directly relevant prior work by
training protocol (one compression level per trained model, paired levels,
or several levels mixed) and by study type; a positive finding under the first two is not a
direct test of the mixed-level setting studied here.

\subsection{Foundation-model trunks}
The detection frontier has moved
substantially toward frozen foundation models with light adaptation:
a linear probe on frozen CLIP features \cite{ojha2023universal} and
forgery-aware adapters over the same trunk \cite{liu2024fatformer} generalize
across unseen generators far better than trained CNNs. Those trunks are stronger, and
orders of magnitude larger, than the 4.0\,M CNN this study uses as its
backbone-matched control, so our verdict reaches them only as a conjecture
(Section~\ref{sec:discussion}). The small-CNN regime nonetheless remains the
one proposed for on-device and high-throughput screening; we measured cost
on a GPU workstation and server only (GPU forward pass on one A100, \SM~Table~\ref{tab:appexpcost};
a detection-to-forward pipeline including its CPU stages, \SM~\S\ref{app:timing}) and make no on-device claim. That regime is also the
one in which the frequency-plus-invariance prescription was formulated and
is still being published.

\subsection{Invariance learning}
The gradient-reversal layer (GRL) originates in
domain-adversarial training \cite{ganin2015unsupervised}; penalties based on
the Hilbert--Schmidt independence criterion (HSIC) \cite{gretton2005measuring} are the standard
non-adversarial alternative. Our registered ablation program included a
GRL-vs-HSIC bake-off; it was correctly never run because the STOP-gate fired
first (Section~\ref{sec:falsification}).

\subsection{Evaluation rigor}
DeepfakeBench \cite{yan2023deepfakebench}
documents how implementation and augmentation details dominate reported
differences between detectors; our augmentation-confound episode
(Section~\ref{sec:confounds}) is a direct instance. We follow
FaceForensics++ \cite{roessler2019faceforensics} protocols and its official
splits, and evaluate on cross-dataset Celeb-DF v2 \cite{li2020celebdf} and
near-domain DFD. Our contribution to this line is procedural: decision
rules registered under git tags \cite{nosek2018preregistration,bertinetto2021preregistration}, and a
structured internal audit of a negative result before publishing it
\cite{karl2024negative}. Internal validation is not external validation \cite{steyerberg2016validation}, which a clinical
study protocol \cite{fernandezcarnero2025frailty} and a clinical prediction study
\cite{fernandezcarnero2023neckpain} both acknowledge by leaving external validation to later work,
and seed and data-sampling variance and data leakage can change a model comparison
\cite{bouthillier2021variance,kapoor2023leakage}; here intra-dataset FaceForensics++ evaluation is
internal, DFD is near-domain rather than cross-dataset, Celeb-DF~v2 is the only genuinely
cross-dataset check outside the exploratory analyses (\SM~\S\ref{app:exploratory}), and the neutral
verdicts rest on three seeds per arm.

\section{CAFRL as specified}\label{sec:cafrl}

This section describes the hypothesis under test, neutrally and as
specified; Sections~\ref{sec:falsification}--\ref{sec:retests} adjudicate
it.

\begin{figure*}[t]
  \centering
  \includegraphics[width=\textwidth]{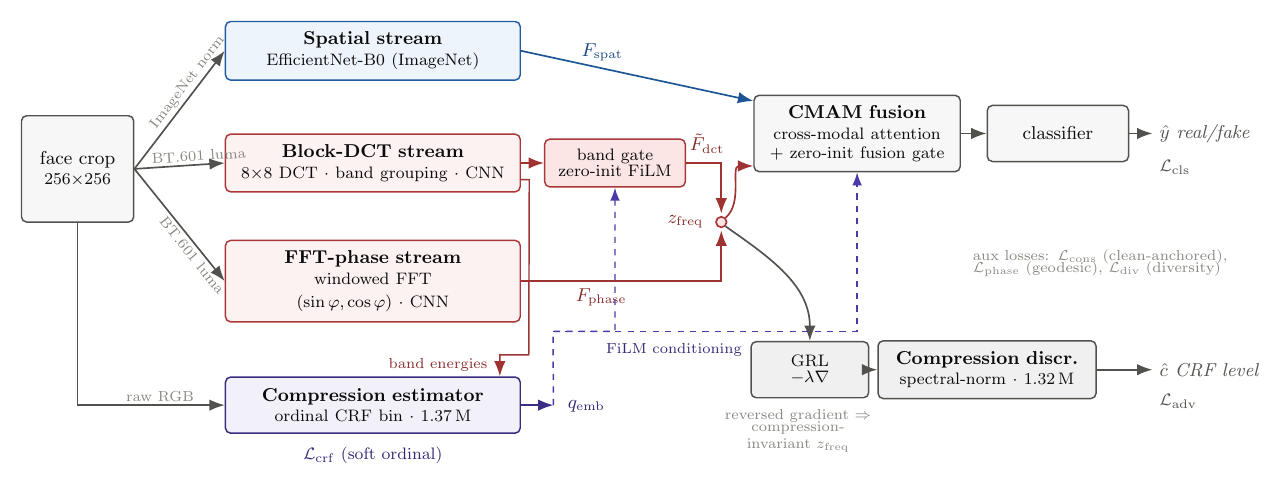}
  \caption{CAFRL as specified. Solid arrows: forward data flow; dashed
  arrows: FiLM conditioning from the compression estimator's embedding
  $q_{\mathrm{emb}}$. The discriminator sees only the frequency representation
  $z_{\mathrm{freq}}$ (the spatial stream is excluded), so the adversarial
  gradient routes to the frequency path alone.}
  \label{fig:arch}
\end{figure*}

\subsection{Design rationale}\label{sec:rationale}
The design was assembled from a structured narrative review, not a
systematic one, run in three rounds in May and June 2026: two rounds of
searches through the Consensus literature search engine --- 28 queries on
novelty, frequency representation and training strategy, which catalogued
the papers that set the macro design, and 58 queries targeting nine
implementation gaps (seven on modules at risk, one on evaluation and one a
currency check), whose roughly 830 candidates
were ranked by relevance to their gap --- and a full-text reading of the
highest-ranked papers that could be obtained. Relevance to a gap was the only
recorded selection criterion, and no second screening or quality appraisal
is recorded. \SM~Table~\ref{tab:rationale} maps each component to the
evidence behind it and marks the choices that were ours
(\SM~\S\ref{app:review}).

\subsection{Architecture}
Fig.~\ref{fig:arch} shows the architecture.
Input is a $256{\times}256$ face crop. Three parallel representations are computed: (i) a \textbf{spatial
stream} --- EfficientNet-B0 \cite{tan2019efficientnet} features
(ImageNet-pretrained); (ii) a \textbf{block-DCT stream} --- $8{\times}8$ block
DCT of the luma channel, band-grouped and encoded by a small CNN; (iii) an
\textbf{FFT-phase stream} --- windowed FFT phase as $(\sin\varphi,
\cos\varphi)$ channels, following evidence that the phase spectrum carries
up-sampling traces \cite{liu2021spsl}. A \textbf{compression estimator} (1.37\,M
parameters) predicts the ordinal CRF bin from the crop; its embedding
conditions a \textbf{frequency-band attention gate} via zero-initialized FiLM
\cite{perez2018film}, so band re-weighting can adapt to the (estimated)
compression level. Streams are fused by cross-modal attention (CMAM
\cite{zhao2023cmam}) with a zero-initialized fusion gate, then classified. A
spectral-normalized \textbf{compression discriminator} (1.32\,M) receives the
frequency representation through a \textbf{gradient-reversal layer}
\cite{ganin2015unsupervised} and enforces compression invariance; per-sample
split BatchNorm separates clean and compressed statistics. Full model:
11.97\,M parameters (10.65\,M without the adversarial branch).

\subsection{Shipped training recipe}
As specified, CAFRL trains 50 epochs
with a three-phase progressive curriculum (clean $\rightarrow$ medium
$\rightarrow$ heavy compression), a frozen backbone for the first 10 epochs,
per-group learning rates (backbone $10^{-5}$, frequency streams $10^{-4}$), a
sigmoid GRL $\lambda$-ramp to 0.5, and auxiliary losses: an L1 clean-anchored
consistency loss and a phase-geodesic loss (weighted by the curriculum 0 and 0
in epochs 1--10, 0.5 and 0.3 in epochs 11--25, and 0.3 and 0.3 from epoch 26); a
circular-relation loss at 0.2 of the phase-geodesic weight; an
attention-diversity loss (1.0); and a soft-ordinal CRF loss (0.2) for the
estimator. Exponential-moving-average (EMA) weights are used for evaluation. The individual
devices --- progressive curricula, consistency and geometry losses, per-group
learning rates --- are common in the compression-robust literature, but the
particular weights, schedule and freeze points are ours;
Sections~\ref{sec:recipeartifact} and~\ref{sec:e13} report what changing this
recipe did.

\textbf{Objective.} Every CAFRL-family arm (CAFRL as specified, and E1,
Fair-N and Fair-W at the fair recipe; \SM~Table~\ref{tab:notation}) minimizes
\begin{equation}
\begin{split}
\mathcal{L} ={}& \mathcal{L}_{\mathrm{cls}}
 + w_{\mathrm{crf}}\mathcal{L}_{\mathrm{crf}}
 + w_{\mathrm{div}}\mathcal{L}_{\mathrm{div}}
 + w_{\mathrm{cons}}(e)\,\mathcal{L}_{\mathrm{cons}}\\
 &+ w_{\mathrm{phase}}(e)\bigl(\mathcal{L}_{\mathrm{geo}}
 + w_{\mathrm{circ}}\mathcal{L}_{\mathrm{circ}}\bigr)\\
 &+ w_{\mathrm{aux}}\mathcal{L}_{\mathrm{aux}}
 + \mathbb{1}_{\mathrm{GRL}}\,\mathcal{L}_{\mathrm{adv}},
\end{split}
\label{eq:objective}
\end{equation}
where $e$ is the epoch, $\mathcal{L}_{\mathrm{cls}}$ the label-smoothed
($\varepsilon{=}0.05$) binary cross-entropy of the detection logit,
$\mathcal{L}_{\mathrm{crf}}$ the estimator's soft-ordinal cross-entropy over
the six CRF bins, $\mathcal{L}_{\mathrm{div}}$ the attention-diversity hinge
on the pairwise cosine similarity of the three pooled stream vectors (margin
0.3), $\mathcal{L}_{\mathrm{cons}}$ the L1 distance of a view's frequency
representation $z_{\mathrm{freq}}$ from that of its clean anchor (the anchor
is not back-propagated), $\mathcal{L}_{\mathrm{geo}}$ and
$\mathcal{L}_{\mathrm{circ}}$ the phase head's geodesic and unit-circle
losses, $\mathcal{L}_{\mathrm{aux}}$ Fair-W's binary cross-entropy on its
standalone frequency logit, and $\mathcal{L}_{\mathrm{adv}}$ the
discriminator's class-weighted cross-entropy on the CRF bin, read through a
gradient-reversal layer of strength $\lambda = 0.5\,\bigl(2/(1+\exp(-10\rho))-1\bigr)$
at training progress $\rho$. As specified, $w_{\mathrm{crf}}{=}0.2$,
$w_{\mathrm{div}}{=}1$ and $w_{\mathrm{circ}}{=}0.2$;
$(w_{\mathrm{cons}}, w_{\mathrm{phase}})$ follow the curriculum schedule above;
and $\rho$, hence $\lambda$, stays 0 until epoch 11. At the fair recipe the only
non-zero terms are $\mathcal{L}_{\mathrm{cls}}$,
$0.2\,\mathcal{L}_{\mathrm{crf}}$ and, where the arm has them,
$\mathcal{L}_{\mathrm{adv}}$ (E1, with $\rho$ counted from the first step) or
$0.2\,\mathcal{L}_{\mathrm{aux}}$ (Fair-W). The reversed adversarial gradient
reaches only the encoder of $z_{\mathrm{freq}}$ (its projection and alignment
layers, the DCT and phase streams, the band gate and the estimator), never
the spatial trunk or the classifier, and $\mathcal{L}_{\mathrm{crf}}$ reaches
only the estimator. \SM~\S\ref{app:objective} gives each term's weight in
each configuration, with the configuration key it comes from, the parameters
each term's gradient reaches, and every tensor shape.

\section{Experimental protocol}\label{sec:protocol}

\subsection{Data}
Throughout, c$X$ denotes H.264 encoding at constant rate factor (CRF) $X$ ---
larger $X$ means heavier compression --- and c0 is the uncompressed original.
Training and intra-dataset testing use FaceForensics++
\cite{roessler2019faceforensics}, with all four core manipulation methods
(Deepfakes, Face2Face, FaceSwap, NeuralTextures) and its official
train/val/test identity splits. FaceShifter \cite{li2020faceshifter} is held out
entirely as a narrowly defined held-out manipulation test (one face-swap
method on FF++ footage), not a general evaluation of generalization. From the c0 masters we produce \textbf{real
H.264 re-encodes} at CRF \{28, 32, 36\} in addition to the distributed
c23/c40, giving six training levels \{c0, c23, c28, c32, c36, c40\} plus an
eval-only c18. Face crops are deterministic re-crops: an SCRFD detection
\cite{guo2022scrfd}, aligned by a similarity warp of its five landmarks; the
largest face per frame is kept by default. In 31 FaceForensics++ originals an
automatic check and inspection found the crop switching between people; the
authors chose the face tracks to crop once, with the same rules for every
split, before any model of this study was trained; DFD tracks were chosen by an automatic
rule. Levels c18--c36 reuse the c0 detections, adding no failures of their
own. Each video gives 32 frames spaced uniformly over its cropped frames, or
all if fewer (never in FaceForensics++; no FaceForensics++, DFD or Celeb-DF~v2 video has zero); the spatial trunk
and the frequency streams read the same crop (\SM~\S\ref{app:facetracks},
Table~\ref{tab:appfacetracks}). Multi-quality (``mq'')
training mixes all six levels per epoch batch. Out-of-distribution evaluation uses
cross-dataset Celeb-DF v2 \cite{li2020celebdf} and near-domain DFD (c23/c40).
Two further corpora, which no model saw in training, enter only the
exploratory analysis added in revision: a fixed subset of the DFDC
\cite{dolhansky2020dfdc} training release and the official test split of
WildDeepfake \cite{zi2020wilddeepfake} (\SM~\S\ref{app:exploratory}).

\subsection{Metrics}
Video-level AUC (per-frame scores mean-pooled per
(video, method); 700 test videos $=$ 140 real $+$ 560 fake per level);
$\auccomp{} =$ mean video-AUC over the heavy levels \{c28, c32, c36, c40\};
c23-anchored drop $= \mathrm{AUC}_{\mathrm{c23}} - \auccomp$, and, post hoc,
the c0-anchored drop $= \mathrm{AUC}_{\mathrm{c0}} - \auccomp$, anchored at the
uncompressed level. Uncertainty:
10,000-replicate video-level bootstrap confidence intervals (CIs) per \emph{cell} (one arm
evaluated at one level on one test set); \emph{paired} deltas draw
one shared video-index vector per replicate so both arms are scored on
identical resamples. Multi-seed arms report the 3-seed mean $\pm$ the 95\%
$t$-band ($4.30\,s/\sqrt{3}$) and \emph{pooled} contrasts (per-video mean
across seeds, then the paired bootstrap).

\subsection{Statistical reporting}
Every interval in this paper is a
\textbf{percentile} bootstrap interval over 10,000 replicates drawn from a
fixed seed, so it reproduces to the digit rather than only in distribution; a
replicate that resamples a single class is redrawn, and in a paired $\Delta$
the redraw is shared, so the two arms are always scored on identical
resamples. $P(\Delta{>}0)$ is the fraction of those replicates in which the
difference is positive --- a description of the same distribution, not a
$p$-value. We call a contrast \emph{significant} when its 95\% paired interval
excludes zero, and we never adjudicate a difference between arms by the
overlap of their marginal intervals, which for correlated arms manufactures
false nulls --- with one exception, flagged where it occurs: the training-data
comparison of Section~\ref{sec:data}, whose synthetic-augmentation arm did not
retain per-video scores, so no paired bootstrap is computable for it. Pooled (3 vs.\ 3) intervals average each video's score across an arm's
three seeds and then resample videos, so they are \emph{conditional on the
three seeds that were run}, and they evaluate a three-member score-level ensemble rather than
any single trained model (\SM~\S\ref{app:statsdetail}). The resampling unit is the evaluated video, which
is not self-evidently the exchangeable one: FaceForensics++ names a
manipulated file \texttt{\{target\}\_\{source\}}. We therefore report
\textbf{five resampling units}, the evaluated video and four identity-cluster
definitions (\SM~\S\ref{app:stats}), and re-run every adjudicated contrast
at all five, and again with the seeds resampled. Conditional on the three
seeds run, no verdict of the twelve original confirmatory contrasts depends on
which of these five units is used. The readings that resampling the seeds as
well changes are reported where they occur (list:
\SM~\S\ref{app:statsdetail}).
For this revision we add a post hoc reading and name its unit the
\emph{primary inferential unit}: the seed first (each arm's seeds resampled
with replacement), then the source-identity cluster, with the mean of the
single-model AUCs as the estimand, so that the interval refers to one model
trained by the procedure rather than to the seeds that were run. Its
two-sided $p$ comes from the same $B = 10{,}000$ replicates as its interval,
$p = (2\min\{\#\Delta^{*}{\le}0,\ \#\Delta^{*}{\ge}0\} + 1)/(B+1)$, an
empirical tail count that cannot fall below $1/(B+1)$; the published seed-fixed,
video-unit values stay printed beside it (Table~\ref{tab:contrasts};
\SM~\S\ref{app:stats} and Table~\ref{tab:pmethods}).

The primary estimand is the mean, over an arm's trained models, of each
model's video-level AUC. Training runs are sampled at arm level: each
bootstrap replicate resamples an arm's seeds with replacement, separately for
each arm. Test videos are paired: all arms are scored on the same videos, and
each replicate shares one resample of source-identity clusters across arms;
other cluster units are sensitivity analyses. Three-seed arms reuse the
integers 42, 1 and 2. At one integer, some arms share the same or a near-same
sampling plan, and some the same initialization in common new layers. The
plan match is deliberate for SpatialBase/E3a and E2a/E3b; no other link has a
recorded purpose. The inferential analysis treats seeds of different arms as
unlinked. Leave-one-seed-out results (Section~\ref{sec:datatests}) are
exploratory.

\subsection{What this design can detect}
The minimum detectable effects below are approximate sensitivity calculations,
conditional on the variability observed in these runs, the test design and
the statistical assumptions of the calculation, not fixed properties of the
design.
At $\alpha{=}0.05$ and 80\% power,
the realized precision of the pooled video-level contrasts corresponds to a
minimum detectable effect of \textbf{about 0.9 AUC points}; treating the seed
as the experimental unit ($k{=}3$ per arm) it is \textbf{1.5 points}, and 1.7
points under the two-level estimator of \SM~\S\ref{app:stats}; we quote
\textbf{1.5--1.7\,pt} where the distinction matters and the conservative 1.7
where one number must serve (per-contrast values:
\SM~\S\ref{app:statsdetail}). Under those
conditions, effects smaller than these would have been detected with
probability below 80\%, in either direction --- which is not the same as ``could not have been
observed''. Every neutral verdict below is bounded by that resolution, and
\SM~\S\ref{app:stats} reports the margin within which the neutral contrasts
can --- and cannot --- be called equivalent.

\subsection{Confirmatory and descriptive contrasts}
Twelve comparisons of
the original analyses are confirmatory: the eight adjudicated contrasts of
Table~\ref{tab:contrasts} and the four held-out-manipulation contrasts of
\SM~\S\ref{app:ood}. Each was adjudicated under a rule fixed before the
comparison it governs was run, and Table~\ref{tab:contrasts} names that rule's tier
row by row. The original training-data comparison (Section~\ref{sec:data}) is
post hoc and sits outside that family; claim~C3 rests on the registered test
T4 instead. The five registered tests of the revision
campaign (T1--T5, Section~\ref{sec:campaign}) are a second, separate
confirmatory family, with Holm's procedure over those five only. Everything else is
labeled where it is reported, with the tiers defined below; the compression
sweeps, the cross-dataset
cells, the mechanism read-outs and the baseline reproductions are
descriptive. Over that family of twelve the
Bonferroni threshold is $\alpha/12 = 4.2\times10^{-3}$, and the three
significant results the paper's claims rest on survive it and Holm's
step-down procedure (normal-approximation $p$ from the bootstrap standard
deviation, so no $1/(B+1)$ floor): the STOP-gate contrast
($p = 2.0\times10^{-4}$), the fair-recipe recovery ($1.4\times10^{-5}$) and the
TrunkCtrl $-$ spatial-residual contrast ($5.5\times10^{-6}$); the original real-CRF-versus-JPEG training
gap ($3.3\times10^{-6}$, an unpaired normal approximation, post hoc, so not
comparable with these paired values) is not a member; \SM~\S\ref{app:stats} gives the family and both
procedures. Non-significant leans are treated identically in both directions:
the same rule that retires the full-trunk variant's $+0.22$\,pt pooled lean
retires the shipped recipe's $+2.2$\,pt held-out-manipulation lean.

\subsection{Evidence tiers}\label{sec:tiers}
Each result's evidence tier is written ``timing, role''; the timing
says what had been seen when its rule was frozen. \emph{Registered} rules
were frozen under an annotated git tag before the comparison they govern
was run. Tag \texttt{phase5-prereg-a1} (commit \texttt{65a8ebb})
froze (i) the \textbf{STOP-gate}: full
CAFRL had to beat the strongest baseline on the paired-$\Delta$ at c40, else
the ablation campaign --- six arm families, \textbf{38 runs} in the frozen plan's budget --- would not run;
and (ii) the \textbf{win-rule} for those ablations --- symmetric, able to
declare the full model the loser, with a registered NULL outcome. CAFRL as
specified had been trained and scored when this tag was made; the baseline
and ablation runs it governs started after it. The gate's
binding sentence reads: ``\textbf{A0 passes iff the 95\% CI of $\Delta$ =
AUC(A0) $-$ AUC(best\_baseline) excludes 0 on the positive side} (paired
$p<0.05$)'', the contrast being the paired video-resampling bootstrap
\texttt{paired\zb\_auc\zb\_delta} at c40 video-AUC with $B{=}10{,}000$ shared
index vectors; ``A0'' is CAFRL as specified; the full frozen text is
released verbatim in the repository's \texttt{preregistration/} directory.
The rule binds a comparison \emph{type} ---
the strongest baseline, adjudicated by a paired-$\Delta$ interval --- not a
named checkpoint: the frozen text says \texttt{best\zb\_baseline}, and the
backbone- and augmentation-matched qualifier is ours, added when the same
document's confound provision produced the control that adjudicated the gate.
Tag \texttt{phase6-prereg} (commit \texttt{40e6b1b}, 2026-10-11) froze the
revision campaign's rules (Section~\ref{sec:campaign}). The original result,
the internal audit and the fair re-tests were known when that campaign was designed:
its rules precede its runs, but its hypotheses were formed after that
evidence. Tag \texttt{phase7-prereg} (commit \texttt{46e4c16}, 2026-11-02)
froze the rule of the backbone-learning-rate replication of Revision~2
(Section~\ref{sec:e13}) before the two replication runs it governs were run
(2026-11-04); that replication is exploratory. All three tags are internal repository history, not a third-party
registry receipt, and git author and committer dates are settable by the
committer; the release offers the frozen texts, the tags and the commit
sequence that places each rule before the comparisons it governed ---
inspectable, but self-attested.
\emph{Post-audit pre-specified} (``pre-specified'' for short) rules were each
journaled before its own run executed, but
written after the audit, when the STOP-gate result, the ablation numbers and
the audit's diagnosis were already known: those of the four fair
re-tests (Section~\ref{sec:retests}), of the held-out-manipulation re-check
(Section~\ref{sec:ood}) and of the confirmation seeds
(Section~\ref{sec:seeds}). Pre-specification fixes the analysis before its data
exist; it does not make the hypothesis a priori. \emph{Post hoc} analyses were chosen after the results they analyze
existed: the $\pm1$\,pt equivalence readings, the resampling-unit and
seed-conditioning sensitivity analyses (\SM~\S\ref{app:stats}), among them
the reading this revision names primary (Section~\ref{sec:protocol}), and the
real-CRF-versus-JPEG training comparison (Section~\ref{sec:data}); they
belong to no family and never replace a registered or pre-specified verdict.
The role says how a result is used: \emph{confirmatory}, a member of the
original twelve or of T1--T5 (above); \emph{secondary}, a campaign
analysis registered as secondary, never confirmatory; \emph{exploratory},
added in revision, outside every confirmatory family, no confirmatory verdict; \emph{descriptive}, no test. Every
verdict is reported at its own rule's tier, and the frozen texts of the
original analyses' rules, at both timings, are released verbatim, so the goalposts could not move after the
numbers arrived, in either direction \cite{simmons2011false} --- the
pre-registration discipline of the empirical sciences
\cite{nosek2018preregistration,bertinetto2021preregistration,hofman2023preregistration}
transplanted into an architecture study.

\subsection{Baselines}
The baselines are Xception \cite{chollet2017xception,
roessler2019faceforensics}, F3-Net \cite{qian2020f3net} and FreqNet
\cite{tan2024freqnet}, each run in \emph{both} protocols --- the literature
protocol (c23-trained) and the mq-control protocol (multi-CRF-trained, matched
to CAFRL's data), six runs in all, every one of them reported at the
compressed levels in \SM~Table~\ref{tab:appbench}, with the complete seven-level
sweep released, plus the
\textbf{backbone-matched, smaller control} --- plain
EfficientNet-B0 (4.0\,M, about the size of CAFRL's own backbone), multi-CRF data,
flip-only augmentation, denoted \textbf{SpatialBase}. Capacity is counted throughout in \textbf{trained
parameters} (\SM~\S\ref{app:notation}). ``Backbone-matched'' means matched to
CAFRL's \emph{backbone family}, not to its total size: the control carries 4.0\,M against CAFRL's
11.97\,M, so it is in fact capacity-\emph{disadvantaged}, which is the
conservative direction for a negative and is why we use it. Baseline recipe: uniform learning rate
$2{\times}10^{-4}$, 30 epochs, batch 32, best-by-validation selection.
Baseline numbers are results of these implemented configurations in our
common environment, not rankings of the original methods: pretraining,
parameter counts and augmentation differ between them, F3-Net is its FAD
branch only, and FreqNet is trained from scratch (\SM~\S\ref{app:baselines}).
Compute: 2$\times$ A100 40\,GB; a CAFRL run is ${\approx}7.5$\,h, a baseline
${\approx}4$\,h. Throughout we call CAFRL's own training procedure
(Section~\ref{sec:cafrl}) the \emph{shipped} recipe and this baseline
procedure the \emph{fair} recipe --- ``fair'' because the re-tests of
Section~\ref{sec:retests} apply it symmetrically to both sides of every
comparison. Table~\ref{tab:config} defines the arms the paper compares other than the
baseline reproductions and names the contrast each comparison estimates and
its selection rule; \SM~Table~\ref{tab:notation} gives the full crosswalk,
the baseline reproductions included (\SM~\S\ref{app:notation}).

\begin{table*}[t]
  \centering
  \caption{Model configuration of the arms the main text compares, and what
  each comparison estimates (baseline reproductions, run ids and where each
  arm is reported: \SM~Table~\ref{tab:notation}). \emph{Trunk}: the
  EfficientNet-B0 feature classified, the full 1280-d head or CAFRL's
  320$\rightarrow$256 tap. \emph{Gate}: the estimator-conditioned band gate.
  \emph{Params}: trained parameters (\SM~Table~\ref{tab:notation}).
  \emph{Levels}: c0 and the real H.264 levels c23--c40, unless stated. \emph{Selection}: the
  checkpoint maximizes mean validation AUC at c23 and the level shown, on
  the weights shown, which are also scored. \emph{Contrast estimated}: what
  differs between the two sides of each comparison in
  Tables~\ref{tab:e13}, \ref{tab:contrasts} and~\ref{tab:confirmatory} and
  of E2b's secondary one, entered on one of its arms. The arms differ in several factors at once, so
  a difference estimates the listed factors jointly; an EMA-scored arm
  against a raw-scored one also changes the training driver (no warm-up and
  label smoothing 0.05 against one warm-up epoch and BCE;
  \SM~\S\S\ref{app:baselines} and~\ref{app:objective}). The shipped-recipe
  arms ran distributed on two GPUs, the others on one; \emph{recipe}
  includes this (\SM~\S\ref{app:campaignarms}). \emph{Matched}: the
  same validation levels and weight type on both sides; \emph{asym.}: not,
  followed by the result under one rule where one is printed. E10: the
  exploratory raw-weight re-selection under the baselines' rule
  (\SM~Table~\ref{tab:appexprawsel}).
  Row~$k$: Table~\ref{tab:contrasts}; T1--T5: Table~\ref{tab:confirmatory}.}
  \label{tab:config}
  \scriptsize
  \setlength{\tabcolsep}{2pt}
  \begin{tabular}{@{}R{0.078}R{0.033}R{0.11}R{0.08}R{0.041}R{0.075}R{0.06}R{0.462}@{}}
    \toprule
    arm & trunk & frequency path, fusion & aux.\ losses; adversary & params (M) &
      recipe; levels & selection & contrast estimated; selection \\
    \midrule
    SpatialBase & full & --- & ---; --- & 4.0 & fair & c40, raw &
      comparator in rows 1, 3--7 and in T3--T5 \\
    TrunkCtrl & tap & --- & ---; --- & 3.68 & fair & c40, raw & vs
      SpatialBase (row 7): trunk tap and parameters jointly; matched
      selection. vs
      spatial-residual (row 8): recipe, frequency path, aux.\ losses,
      adversary and parameters jointly; asym.\ (raw vs EMA); one rule: E10,
      points only, no interval (\SM~Table~\ref{tab:appexprawsel}) \\
    CAFRL as specified & tap & DCT, phase, gate, CMAM & as specified; GRL &
      11.97 & shipped & c40, EMA & SpatialBase vs it (row 1, STOP-gate):
      frequency path, fusion, aux.\ losses, adversary, trunk tap, parameters
      and recipe jointly; asym.\ (raw vs EMA); one rule: E10, 3-seed-mean
      points only, no interval (\SM~Table~\ref{tab:appexprawsel}) \\
    CAFRL$-$adv & tap & as CAFRL & as specified; --- & 10.65 & shipped &
      c40, EMA & comparator in row 2 \\
    spatial-residual & tap & as CAFRL; logit $=$ spatial
      $+\,\gamma\cdot$frequency & as specified; GRL & 10.29 & shipped & c40,
      EMA & comparator in row 8 \\
    Fair-N & tap & as CAFRL & CRF (0.2); --- & 10.65 & fair & c40, EMA & vs
      CAFRL$-$adv (row 2): the recipe as a whole, one architecture; matched.
      vs SpatialBase (rows 3, 5): frequency path, CMAM, aux.\ loss, trunk tap
      and parameters jointly; asym.\ (EMA vs raw); one rule: E10, three
      seeds, $-0.37$\,pt
      (Section~\ref{sec:seeds}). vs TrunkCtrl (post hoc row): frequency path,
      CMAM, aux.\ loss and parameters jointly; asym.; no one-rule contrast printed \\
    Fair-W & full & DCT, phase, gate; $\gamma$-residual & CRF, frequency
      head (0.2 each); --- & 9.63 & fair & c40, EMA & vs SpatialBase (rows 4, 6):
      frequency path, $\gamma$-residual fusion, aux.\ losses and parameters jointly, one trunk; asym.\
      (EMA vs raw); one rule: E10, three seeds, $-0.72$\,pt (Section~\ref{sec:seeds}) \\
    E1 & tap & as CAFRL & CRF (0.2); GRL & 11.97 & fair & c40, EMA & vs Fair-N
      (T1, T2): the adversary only; matched; raw-weight re-selection
      0.06\,pt (Section~\ref{sec:adversary}) \\
    E2a & full & --- & ---; --- & 4.0 & fair; c0--c36 & c36, raw & with
      anchor and SpatialBase (T3): retention of the c40 gain without c40 in
      training; with E3a, E3b (T5): c40-level data in training, averaged over
      codec; both asym.\ (c36 vs c40); symmetric-selection control,
      SpatialBase re-selected by E2a's rule (anchor, E3a keep c40): $R$ 0.94, T5 1.58\,pt
      (Section~\ref{sec:datatests}; \SM~Table~\ref{tab:appcampaignsec}) \\
    E2b & full & --- & ---; --- & 4.0 & fair; c0--c32 & c32, raw & vs E2a
      and SpatialBase (secondary): the heaviest training levels; asym.; no
      one-rule contrast printed \\
    E3a / E3b & full & --- & ---; --- & 4.0 & fair; c0 $+$ JPEG & c40 / c36,
      raw & vs SpatialBase / E2a (T4): real H.264 against PSNR-matched JPEG
      training levels (E3a c23--c40, E3b c23--c36; validation on H.264),
      averaged over c40 exposure; matched within each pair \\
    anchor & full & --- & ---; --- & 4.0 & fair; c23 & c40, raw & zero point
      of T3's ratio; not tested on its own \\
    E13 (five arms) & tap & as CAFRL & CRF (0.2), or set back; --- & 10.65 & fair,
      one change set back & c40, EMA & vs Fair-N's seed-42 run
      (Table~\ref{tab:e13}): one shipped-recipe change reverted at a time
      (the curriculum arm with its own loss weights), one seed; matched;
      the backbone-LR arm also at seeds 1, 2 vs Fair-N's seeds 1, 2 (rule:
      \texttt{phase7-prereg}; Section~\ref{sec:e13}) \\
    \bottomrule
  \end{tabular}
\end{table*}

\subsection{The revision campaign}\label{sec:campaign}
For the first revision we trained 23 new runs (\SM~Table~\ref{tab:notation}), each on
one GPU at a batch of 32, and scored each on every test set above, keeping
its per-video predictions; Revision~2 added two (Section~\ref{sec:e13}). The arms (Table~\ref{tab:config}; definitions,
E1's adversary ramp and the recipe differences left unreverted:
\SM~\S\ref{app:campaignarms}) are \emph{E1}, Fair-N with the adversarial
compression-invariance loss restored; \emph{E2a} and \emph{E2b}, SpatialBase
trained on c0--c36 and on c0--c32 (E2b descriptive only); \emph{E3a} and
\emph{E3b}, SpatialBase and E2a with PSNR-matched JPEG in place of the real
H.264 levels; the \emph{anchor}, SpatialBase on c23 only, the zero point of
T3's retention ratio; and \emph{E13} (one run per arm, seed 42;
descriptive), Fair-N with one recipe change reverted to the shipped value
(Section~\ref{sec:e13}).

\textbf{Registration.} The campaign's rules --- the five confirmatory
tests T1--T5 with Holm's procedure over them (Table~\ref{tab:confirmatory}),
the primary estimand and its hierarchical bootstrap, the minimum detectable
effects, the secondary analyses, the run-health and stop rules
(\SM~\S\ref{app:campaignarms}), and the
sentence that reports each possible outcome --- were written and checked,
approved on 2026-10-11 and frozen by the tag \texttt{phase6-prereg} before
the first campaign run was launched; they are
registered, with what was known and the self-attestation caveat as
Section~\ref{sec:tiers} states.

\textbf{Symmetric selection.} A registered secondary analysis removes one
asymmetry of selection: E2a, E2b and E3b, trained without c40, cannot select
on c40, while SpatialBase and the anchor select on c23 and c40
(Table~\ref{tab:config}). SpatialBase's three runs are re-selected from their
stored per-epoch checkpoints by E2a's rule, and T3 and T5 are recomputed with
these runs and reported beside the registered results, never in their place
(Section~\ref{sec:datatests}; the rule and the epochs it picks:
\SM~\S\ref{app:campaignarms}).

\section{Results I --- the initial configuration fails its advancement criterion}\label{sec:falsification}

\subsection{The STOP-gate fires}\label{sec:stopgate}

\begin{figure*}[t]
  \centering
  \includegraphics[width=\textwidth]{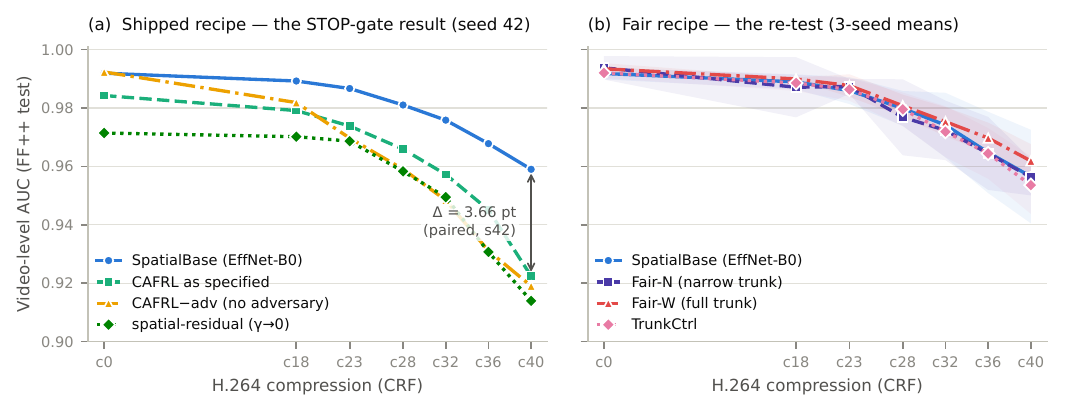}
  \caption{Video-AUC across compression levels (FF++ test, 700 videos per
  level). (a)~The STOP-gate result at the shipped recipe, \emph{seed 42} --- the arms
  as the STOP-gate adjudicated them, matching Table~\ref{tab:stopgate}
  and \SM~Table~\ref{tab:sweep} cell for cell; the annotated c40 gap is the adjudicated
  contrast $+0.0366$, paired on test videos, a configuration result (as
  configured, one seed), not an architectural effect (the two further shipped-CAFRL seeds lie within
  0.002 of the plotted curve at c40). (b)~The fair-recipe re-tests, plotted as
  \emph{3-seed means} with 95\% $t$-bands ($4.30\,s/\sqrt{3}$, $k{=}3$), so
  its SpatialBase point at c40
  is the 3-seed mean 0.9565 rather than \SM~Table~\ref{tab:sweep}'s seed-42 value
  0.9590. \SM~\S\ref{app:seeds} prints the per-seed values at c40 and
  \auccomp{}; panel~(b)'s remaining levels are 3-seed means that appear in no
  printed cell and are in the release, the one place this paper departs from
  its rule that every plotted point is a printed one.}
  \label{fig:sweep}
\end{figure*}

Table~\ref{tab:stopgate} (rows 1--2) and Fig.~\ref{fig:sweep}a give the gate's
comparison, with a backbone-matched, smaller control, matched data and matched
augmentation for the two arms it adjudicated (the Xception-mq row is shown for
context and uses the standard augmentation policy; see Section~\ref{sec:confounds}).

\begin{table}[t]
  \centering
  \caption{The STOP-gate comparison (video-AUC, FF++ test). Every row is a
  single run at seed 42 --- the arms exactly as the registered gate
  adjudicated them. Xception-mq is our implemented configuration, not a
  ranking of the original method (Section~\ref{sec:protocol}).}
  \label{tab:stopgate}
  \footnotesize
  \setlength{\tabcolsep}{3pt}
  \begin{tabular}{@{}lcccc@{}}
    \toprule
    model & params & c23 & c40 & \auccomp \\
    \midrule
        \textbf{SpatialBase} & \textbf{4.0\,M} & \textbf{.9867} & \textbf{.9590} & \textbf{.9709} \\
    CAFRL as specified & 11.97\,M & .9738 & .9224 & .9477 \\
    Xception-mq & 20.8\,M & .9872 & .9679 & .9765 \\
    \bottomrule
  \end{tabular}
\end{table}

Under FF++-derived single-pass H.264 re-encoding, CAFRL's curve is flat and
robust by literature standards (c23-anchored drop 0.026) --- but the plain backbone's is flatter, and higher at every compressed
level. (At c0 the adversary-free variant CAFRL$-$adv is 0.04\,pt above it, a
difference far inside this design's ${\approx}0.9$\,pt resolution;
\SM~Table~\ref{tab:sweep}.) The paired STOP-gate contrast $\Delta$(SpatialBase $-$ CAFRL as specified) at
c40 is $\mathbf{+0.0366}$, 95\% CI \ci{+0.020}{+0.054} --- both arms at seed
42, 10{,}000-replicate paired video bootstrap --- significant, and
confirmatory under the registered rule quoted in
Section~\ref{sec:protocol}, with roughly one-third the parameters
(4.0\,M vs.\ 11.97\,M) and none of the frequency machinery. Per the registered rule, the ablation campaign did not run.

\subsection{Confound discipline}\label{sec:confounds}

Two alternative explanations were caught during the campaign and controlled
before concluding:

\begin{itemize}
\item \textbf{Augmentation.} A stronger augmentation policy (added to
reproduce literature baseline anchors) accidentally advantaged the mq-control
baselines over flip-only CAFRL. Re-running the controls flip-only changed
little on mq data (Xception-mq $0.968 \rightarrow 0.963$ at c40): augmentation
was \emph{not} the explanation. (It mattered greatly for c23-trained
literature baselines --- $+3$ to $+6$ frame-AUC points at c40 --- consistent
with \cite{yan2023deepfakebench}; details in \SM~\S\ref{app:baselines}.)
\item \textbf{Capacity.} Xception's win could reflect its 20.8\,M parameters.
The backbone-matched, smaller SpatialBase control (4.0\,M) still beat CAFRL --- capacity was not
the explanation either.
\end{itemize}

\subsection{The causal chain, as originally read}\label{sec:originalchain}

With the recipe held fixed at CAFRL's own, ablations traced the deficit (all
at c40): removing the GRL did not recover the deficit (CAFRL$-$adv 0.919 vs.\ full 0.922 ---
the adversary is not the culprit here); a do-no-harm spatial-residual redesign (logit $=$
spatial $+$ $\gamma\cdot$frequency, $\gamma$ zero-initialized) also did not
recover it (0.914) and drove $\gamma \rightarrow 0$. At the time we read this as
``the model freely discards frequency; frequency is unwanted; the spatial path
is architecturally bottlenecked.'' \textbf{Section~\ref{sec:audit} shows this
reading was wrong} --- the measurement stands, the mechanism does not.

This is also the evidence for what this paper says about the adversary at the
shipped recipe (with D6, Section~\ref{sec:audit}). The fair re-tests of
Section~\ref{sec:retests} all run with the adversary off, to give the
frequency \emph{representation} its best chance, not to re-tune the
adversary; the revision campaign tested the adversary at the fair recipe
(Section~\ref{sec:adversary}).

\section{The internal audit}\label{sec:audit}

The campaign flow is drawn in \SM~Fig.~\ref{fig:flow}.

Before writing the negative up, we audited it as an adversary would --- four
author-directed, tool-assisted internal audit passes (training parity, the
frequency-path implementation, evaluation symmetry and the run diagnostics);
they are internal quality control, not external validation, and can share
the authors' assumptions and errors (their procedure:
\SM~\S\ref{app:auditproc}). The \textbf{measurement survived} (every
video-level AUC quoted at the time recomputed exactly from the stored per-video scores;
evaluation fully symmetric; no leakage;
selection on validation only). The \textbf{causal interpretation did not.}
The audit found eight defects, D1--D8, of which four biased the comparison \emph{against}
frequency: \textbf{D1}, an L1 consistency loss running 40--120$\times$ the
classification loss in magnitude (a ratio of logged loss values; gradient
norms were not recorded) at the curriculum's compressed-phase onset, in
\emph{every} arm including CAFRL$-$adv; \textbf{D2}, the residual probe's
$\gamma \rightarrow 0$, a shock collapse locked in by a structural gradient
deadlock rather than a choice; \textbf{D3}, an optimization that was never
matched --- a 20$\times$ backbone learning-rate gap, a 10-epoch freeze, heavy
data deferred to the learning-rate tail; and \textbf{D4}, a spatial path
tapping an intermediate 320-channel map and discarding the backbone's final
1280-d representation. Each is repaired by a named re-test below (D1 by
Fair-N, D2 by Fair-W, D3 by TrunkCtrl and Fair-N, D4 by TrunkCtrl and
Fair-W; Table~\ref{tab:retests}).
Appendix~\ref{app:ledger} is the authoritative telling: the complete D1--D8
ledger, why each defect mattered, and what closed it.

The lesser defects D5--D8 are in Table~\ref{tab:appledger}, and the
training-health guardrails that now ship in the codebase in
\SM~\S\ref{app:auditproc}.

\section{Results II --- the fair re-tests}\label{sec:retests}

Four re-tests, each with its decision rule journaled before the run
(Table~\ref{tab:retests}): the \textbf{routing probe}, an evaluation-only
check of the split-BatchNorm routing (D5) that re-scores existing checkpoints
and is not a new training configuration; \textbf{TrunkCtrl}, the trunk-recipe control; \textbf{Fair-N}, the fair
frequency test at CAFRL's narrow trunk; and \textbf{Fair-W}, the same test at
the full trunk (\SM~Table~\ref{tab:notation}; Fig.~\ref{fig:sweep}b). The
held-out-manipulation re-check (Section~\ref{sec:ood}) and the confirmation
seeds (Section~\ref{sec:seeds}) are not re-tests.

\begin{table}[t]
  \centering
  \caption{The four post-audit pre-specified re-tests: each one's decision
  rule (journaled before its run, one phrase here), the defects it addresses
  (Section~\ref{sec:audit}) and its principal outcome at c40. The last two
  rows are not re-tests and are listed only to keep them apart. {\mathversion{normal}$\Delta$}:
  point estimate.}
  \label{tab:retests}
  \footnotesize
  \setlength{\tabcolsep}{2pt}
  \begin{tabular}{@{}>{\raggedright\arraybackslash}p{0.18\columnwidth}>{\raggedright\arraybackslash}p{0.34\columnwidth}>{\raggedright\arraybackslash}p{0.09\columnwidth}>{\raggedright\arraybackslash}p{0.32\columnwidth}@{}}
    \toprule
    re-test & decision rule & defects & principal outcome \\
    \midrule
    routing probe (eval only) & material iff oracle routing raises c40
      AUC by ${\ge}0.005$ for CAFRL as specified or CAFRL$-$adv & D5 &
      immaterial: $\Delta \le 0$ on both arms \\
    \addlinespace[1.5pt]
    TrunkCtrl (3 seeds) & recipe-dominant iff $\Delta$(SpatialBase $-$
      TrunkCtrl) CI includes 0 or TrunkCtrl (seed 42) ${\ge}0.949$ & D3, D4 &
      recipe-dominant: TrunkCtrl 0.9536; $\Delta$ $-0.0038$
      \ci{-0.0116}{+0.0039} \\
    \addlinespace[1.5pt]
    Fair-N & $\Delta$(Fair-N $-$ SpatialBase): CI excludes 0 for or against
      Fair-N, else neutral & D1, D3 &
      neutral: $-0.0004$ \ci{-0.0092}{+0.0080}; $+0.0396$ over CAFRL$-$adv \\
    \addlinespace[1.5pt]
    Fair-W & the same three-way rule for $\Delta$(Fair-W $-$ SpatialBase) &
      D2, D4 & neutral: $+0.0055$ \ci{-0.0022}{+0.0134}; $\gamma$ deadlock
      broken \\
    \midrule
    FaceShifter re-check (not a re-test) & a claim only if the pooled
      paired-$\Delta$ CI excludes 0 in the favorable direction & --- &
      no claim (Section~\ref{sec:ood}) \\
    \addlinespace[1.5pt]
    confirmation seeds (not a re-test) & pre-stated quoting and STOP rule;
      descriptive: 3 vs.\ 3 pooled contrasts
      against the SpatialBase pool & --- & both CIs include 0
      (Table~\ref{tab:contrasts}) \\
    \bottomrule
  \end{tabular}
\end{table}

\subsection{Changing the training recipe removed the deficit}\label{sec:recipeartifact}

\textbf{TrunkCtrl (architecture vs.\ recipe, 3 seeds):} CAFRL's exact 320$\rightarrow$256
spatial trunk alone, trained with the plain baseline recipe, reaches c40
seed-mean $0.9536$ (3 seeds) against SpatialBase's $0.9590$ (seed 42). The
gap between SpatialBase and the shipped-recipe spatial-residual arm at c40 is $0.9590
- 0.9139 = 0.0451$, both single runs at seed 42, and TrunkCtrl sits between
the two:
\[
  \underbrace{0.0451}_{\text{SpatialBase}\,-\,\text{sp.-res.}}
  \;=\;\underbrace{0.0397}_{\text{TrunkCtrl}\,-\,\text{sp.-res.}}
  \;+\;\underbrace{0.0054}_{\text{SpatialBase}\,-\,\text{TrunkCtrl}},
\]
where ``sp.-res.'' is the shipped-recipe spatial-residual arm and TrunkCtrl
enters as its 3-seed mean. These are descriptive differences, not shares of
cause: TrunkCtrl also lacks the frequency stream, auxiliary losses and adversary
the spatial-residual arm carries, has fewer parameters and is scored on raw rather
than EMA weights, so TrunkCtrl $-$ sp.-res.\ is not a pure recipe effect, and the change of recipe alters loss weights, curriculum, learning rates,
freezing, training length and GPU count at once, so the contributions of its individual modifications are
not identified. The paired bootstrap supplies each difference's
uncertainty --- $\Delta$(SpatialBase $-$ TrunkCtrl) $-0.0038$, CI
\ci{-0.0116}{+0.0039}, no detectable difference between the two implemented
spatial controls; $\Delta$(TrunkCtrl $-$ spatial-residual) $+0.0489$, CI \ci{+0.0294}{+0.0705},
significant --- with the caveat that both bootstrap contrasts average TrunkCtrl's
per-video scores over its three seeds before pairing, a different estimand
from the single-run point differences above. This revision's post hoc
primary reading takes TrunkCtrl's mean single-model AUC instead and returns
the point differences above, $+0.0054$ \ci{-0.0040}{+0.0156} and $+0.0397$
\ci{+0.0196}{+0.0625} (EMA for spatial-residual, raw for SpatialBase and
TrunkCtrl; Table~\ref{tab:contrasts}): the sign of the first and
the smaller size of the second come mostly from the change of estimand, not
from resampling the seeds.
The two spatial controls differ in more than the width of the representation
they classify from --- in their processing and in parameter count (4.0\,M
against 3.68\,M) --- so their tie does not isolate the cost of tapping the
320-channel map. Both estimands give the same reading, the
rule's verdict {\bfseries\boldmath recipe-dominant}: the revised training
configuration eliminated most of the trunk gap (under both, the only interval
that excludes zero is TrunkCtrl's over the shipped-recipe arm);
Section~\ref{sec:e13} reverts its modifications one at a time.

\textbf{Fair-N (the fair frequency test):} the identical full frequency stack ---
DCT and phase streams, estimator-conditioned band gate, CMAM fusion --- with the
adversary off, \emph{zero} invariance/geometry losses, flat curriculum,
uniform learning rate $2{\times}10^{-4}$, 30 epochs. Result: c40 \textbf{0.9586 vs.\ SpatialBase
0.9590} --- $\Delta = -0.0004$, CI \ci{-0.0092}{+0.0080}: no detectable
difference for this implementation, trained without a hyperparameter search of
its own (Limitation~(iii); its 3-seed form: Section~\ref{sec:seeds}). Against
the shipped-recipe counterpart (CAFRL$-$adv, 0.9190, seed 42), the fair recipe,
as a package on the same components, bought $\mathbf{+0.0396}$, CI \ci{+0.0231}{+0.0570}
--- significant; which of its changes did so is not identified.
These are two different contrasts and should be read as such: the STOP-gate
deficit is SpatialBase against CAFRL as specified ($+0.0366$), while the recovery is Fair-N
against CAFRL$-$adv ($+0.0396$), the shipped-recipe arm built from Fair-N's
own components. Read together they place the ``machinery hurts'' deficit in
the recipe package as a whole rather than in the frequency content; they do
not show that every component is irrelevant.

\subsection{One recipe change at a time (registered,
descriptive)}\label{sec:e13}

\begin{table}[t]
  \centering
  \caption{The registered one-seed recipe ablation E13 (tier: registered, descriptive; no
  test, no $p$-value, no verdict). Each row is Fair-N with one change reverted
  to its value in the shipped recipe, minus Fair-N's seed-42 run: AUC points
  on the FF++ test split, with fixed-seed 95\% percentile intervals (only the
  140 source-identity clusters are resampled, $B = 10{,}000$; one run per
  arm, seed 42). {\mathversion{normal}$\Delta$}: point estimate. Weights: EMA,
  selected by Fair-N's rule. Fair-N's seed-42 run scores 95.86 at c40 and 97.14 on \auccomp.
  The backbone-learning-rate row replicated on two new seeds, with the same
  sign convention (exploratory; Section~\ref{sec:e13}).}
  \label{tab:e13}
  \footnotesize
  \setlength{\tabcolsep}{2pt}
  \begin{tabular}{@{}>{\raggedright\arraybackslash}p{0.33\columnwidth}ll@{}}
    \toprule
    change reverted & {\mathversion{normal}$\Delta$} c40 & {\mathversion{normal}$\Delta$} \auccomp \\
    \midrule
    training curriculum (identical re-run, epoch 29)
      & $-0.25$ \ci{-1.16}{0.68} & 0.004 \ci{-0.62}{0.67} \\
    backbone learning rate & $-4.07$ \ci{-5.80}{-2.45} & $-2.71$ \ci{-3.97}{-1.58} \\
    frequency-stream learning rate & 0.14 \ci{-1.00}{1.33} & $-0.08$ \ci{-0.82}{0.70} \\
    auxiliary-loss weights & 0.10 \ci{-0.94}{1.18} & 0.30 \ci{-0.31}{1.01} \\
    training length & 0.05 \ci{-1.02}{1.05} & $-0.13$ \ci{-0.93}{0.65} \\
    \bottomrule
  \end{tabular}
\end{table}

The comparison above changes the whole recipe at once; the registered
ablation E13 (Section~\ref{sec:campaign}) changes one thing at a time
(Table~\ref{tab:e13}). The registered sentence for each arm, with its
numbers filled in:

In a one-seed ablation, reverting the training curriculum (flat back to the
shipped three-phase progressive curriculum, phase bounds 10 and 25) changed c40 AUC by
$-0.25$\,pt against Fair-N (fixed-seed interval \ci{-1.16}{0.68};
descriptive). In a one-seed ablation, reverting the backbone learning rate
($2{\times}10^{-4}$ back to $10^{-5}$) changed c40 AUC by $-4.07$\,pt against
Fair-N (fixed-seed interval \ci{-5.80}{-2.45}; descriptive). In a one-seed
ablation, reverting the frequency-stream learning rate ($2{\times}10^{-4}$
back to $10^{-4}$) changed c40 AUC by 0.14\,pt against Fair-N (fixed-seed
interval \ci{-1.00}{1.33}; descriptive). In a one-seed ablation, reverting
the auxiliary-loss weights (zero back to consistency 0.5, phase-geodesic 0.3,
circular-relation 0.2 and attention-diversity 1.0) changed c40 AUC by
0.10\,pt against Fair-N (fixed-seed interval \ci{-0.94}{1.18}; descriptive).
In a one-seed ablation, reverting the training length (30 epochs back to 50,
with the learning-rate schedule stretched to match) changed c40 AUC by 0.05\,pt against
Fair-N (fixed-seed interval \ci{-1.02}{1.05}; descriptive).

The curriculum arm carries a confound stated in the registration. With the
literal phase bounds inside Fair-N's 30 epochs, c36 and c40 are seen only in
epochs 26--30, at no more than 7\% of the peak learning rate; and the
progressive curriculum applies its own hard-coded consistency and
phase-geodesic weights (0 and 0 in epochs 1--10, with the backbone frozen;
0.5 and 0.3 in epochs 11--25; 0.3 and 0.3 from epoch 26), so the arm reverts
the curriculum together with those weights and cannot separate them.
The first run of this arm diverged at epoch 14; we re-ran it with nothing
changed and report the re-run, a deviation from the registered plan that
rests on one identical re-run, not three, because a three-run check was too expensive
(\SM~\S\ref{app:curriculum}). The registered outcome against our reading,
that no single reverted change moves c40 AUC or \auccomp{} outside Fair-N's
fixed-seed interval, did not occur: on the re-run, only the backbone learning
rate moves both endpoints.

\textbf{Replication of the backbone-learning-rate arm (exploratory).} In
Revision~2 this arm was run at seeds 1 and 2 and compared with Fair-N's
seeds 1 and 2, under a rule frozen under the tag \texttt{phase7-prereg}
before the two new runs were run (Section~\ref{sec:tiers}): the result counts as replicated if the 95\%
interval of reverted minus Fair-N at c40, from seeds 1 and 2 of each arm
only, excludes zero on the side of the seed-42 result. It replicated:
$-3.06$\,pt \ci{-4.85}{-1.42} (arm means of single-model AUC over seeds 1
and 2, seeds then source-identity clusters resampled, $B = 10{,}000$; no
$p$-value). Beside it, not decisive: $-3.39$\,pt \ci{-5.12}{-1.78} over all
three seeds. This remains a sensitivity of one recipe factor, not a causal
decomposition of the recipe effect: the other changes were not replicated,
and the changes were not tested in combination.

\subsection{The frequency sub-network is discriminative on its own but adds
no detectable marginal value in these implementations}\label{sec:redundancy}

\textbf{Fair-W (frequency at full strength):} the backbone's full 1280-d
representation drives the primary logit (repairing D4); the frequency
representation enters as a zero-initialized $\gamma$-residual; an auxiliary
binary cross-entropy (BCE) on the standalone frequency logit supplies label gradient regardless of
$\gamma$ (repairing D2). Result: c40 \textbf{0.9645 vs.\ SpatialBase 0.9590} --- $\Delta
= +0.0055$, CI \ci{-0.0022}{+0.0134}, $P(\Delta{>}0)=0.93$: a lean, not a
win --- descriptive, within two standard errors of zero, and smaller than the
effect this design could resolve. It is the first CAFRL variant at or above
SpatialBase at every level, and no rule in this paper licenses a claim from
it.

\begin{figure}[t]
  \centering
  \includegraphics[width=0.95\columnwidth]{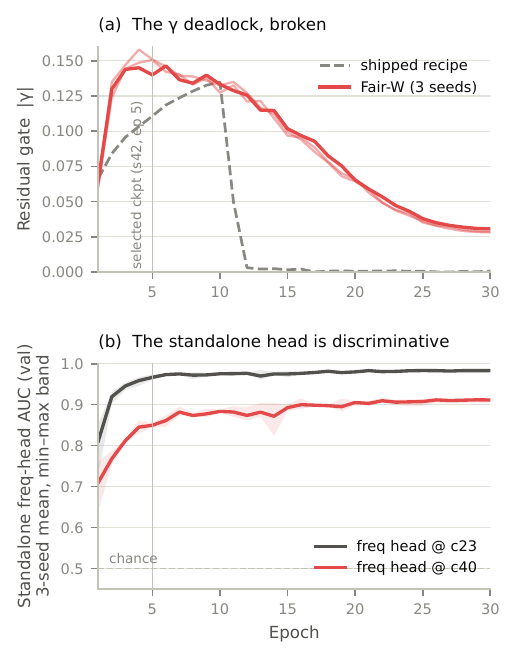}
  \caption{Frequency-path read-outs (Fair-W; all three seeds, seed 42 in the
  foreground). (a)~The $\gamma$ deadlock is broken: the residual weight grows
  to 0.146 by epoch 6 and decays to 0.031 by epoch 30, against the
  shipped-recipe probe (dashed), which falls from 0.135 to 0.003 across the
  curriculum shock at epochs 11--12. The vertical line is the selected
  checkpoint (seed 42, epoch 5, $\gamma = 0.140$; seeds 1 and 2 select 6
  and 7). (b)~The standalone frequency head (3-seed mean; band $=$ per-epoch
  min--max) plateaus at ${\approx}0.91$/${\approx}0.98$ (c40/c23) validation
  AUC over the second half of training, from 0.71/0.81 (c40/c23) at epoch 1 and 0.85/0.97 at the selected
  checkpoint: its output is discriminative, including at c40, yet adds no
  detectable value over the trunk in this implementation. What it exploits was not isolated to frequency --- no matched
  pixel-domain control was trained.}
  \label{fig:mechanism}
\end{figure}

The mechanism read-outs (Fig.~\ref{fig:mechanism}) exclude one explanation
of the tie but do not settle why it occurs. The $\gamma$-deadlock is
genuinely broken (Fig.~\ref{fig:mechanism}a): $\gamma$ declines after its
early peak but never collapses as the shipped-recipe probe's does (seeds 1
and 2 peak at 0.158 and 0.151, the same shape as seed 42). And the
\emph{frequency-stream sub-network is individually discriminative}
(Fig.~\ref{fig:mechanism}b). The model uses the frequency path, and the frequency sub-network's own
output is discriminative --- and conditional on a strong spatial trunk trained
on diverse-compression data, its marginal value under this fusion is
${\approx}0$. One explanation is thereby excluded: the head reaches
${\approx}0.91$ validation AUC \emph{at c40}, so the tie is not the tie of a
signal that CRF-40 encoding has destroyed. What remains undecided is whether
the trunk already carries the information or the fusion never let it speak;
the exploratory test-split analyses below bear on that question without
deciding it. \textbf{No detectable marginal value under
this fusion, for these implementations and without a hyperparameter search
--- not shown to be absent.}

The conditioned band gate is not idle either. Zero-initialized to the
identity ($g \equiv 1$), it leaves that initialization at the first logged
epoch and never returns: on the c23 validation diagnostic the per-band mean
gate at the selected checkpoint spans $0.25$--$1.61$ in Fair-N and
$1.37$--$13.6$ in Fair-W (seed 42), with all three Fair-W seeds learning the
same high-band emphasis, roughly an order of magnitude above the low band. The
component is therefore exercised in the arms that carry the neutral verdict,
which is what the verdict needs. The training diagnostic is a c23-only
pass, a trajectory over epochs at one compression level; an exploratory
test-split read-out of the gate at all seven levels, which also relates the
estimator's errors to detection (EMA weights; \SM~Tables~\ref{tab:appexpbandgate}
and~\ref{tab:appexpestimator}), is descriptive and does not separate
conditioning on the estimated compression level, the property the gate's
design claims, from a response to the input itself.

Two limits belong here, where the claim is made rather than only in the
Limitations. First, this is a \emph{fusion} result: it shows that adding
these streams to this trunk through CMAM or a $\gamma$-residual, as
implemented here and without a hyperparameter search, buys nothing
measurable, not that the information they carry is missing from the trunk's
own representation. The stored scores held only the fused logit, so for this
revision a fresh forward pass on the test split supplied three read-outs
(exploratory, no verdict; EMA weights for Fair-W, raw for SpatialBase;
\SM~\S\ref{app:exploratory}, \SM~Table~\ref{tab:appexpdiag}). The arm-invariant
selector lands Fair-W at epoch 5 of 30 (6 and 7 at the other seeds), where
$\gamma$ peaks; scored alone there, the standalone head reaches 85.27 AUC
points at c40, and 89.99 at epoch 30 (the fused Fair-W model: 96.19 and 94.66). Its
share of the fused logit, $|\gamma f| / (|s| + |\gamma f|)$ with $s$ the
spatial and $f$ the head logit, averages 0.11 at c40. Added to SpatialBase's
logit at unit weight, with no fitted parameter, the head changes
SpatialBase's c40 AUC by 0.01\,pt \ci{-0.99}{1.02}, and it errs on 55\%
of the videos SpatialBase errs on (Jaccard index of the two error sets 0.23).
These read-outs do not separate the two readings above: a unit-weight score
fusion is one further fusion, not a test of what the trunk already carries.
Second, the standalone head shows that the
\emph{sub-network} is discriminative, not that what it exploits is
specifically frequency: no matched-capacity pixel-domain CNN was trained on
the same crops as a control, and the training diagnostics log the combined
head only, so the block-DCT and FFT-phase contributions are not separated
either.

\subsection{Seed confirmation (3 vs.\ 3)}\label{sec:seeds}

Fair-N and Fair-W were re-run at two additional seeds (configs differing only in the
seed), making every contrast against the 3-seed SpatialBase pool symmetric
(Table~\ref{tab:contrasts}; per-seed values in \SM~\S\ref{app:seeds}).

\begin{table*}[t]
  \centering
  \caption{The adjudicated contrasts at c40 (10{,}000-replicate paired video
  bootstrap, 700 videos). The first eight rows are confirmatory, at the evidence
  tier the \emph{rule} column names (Section~\ref{sec:tiers}).
  ``s42'' is that arm's frozen seed-42 run; ``pool'' averages an arm's
  per-video scores over its three seeds before the bootstrap, so pooled
  intervals are conditional on those seeds. ``No detectable difference'' is
  bounded by this design's resolution (Section~\ref{sec:protocol}).
  Rows 7--8 are deliberately asymmetric, as the trunk-recipe control's
  pre-specified rule required; pooling SpatialBase's seeds there moves the
  SpatialBase $-$ TrunkCtrl estimate to about $+0.002$ and changes no verdict.
  \SM~\S\ref{app:stats} re-runs every row with the seeds resampled and at
  four cluster-level units.
  The last column and the last row are post hoc (Revision 1) and replace no
  published value. {\mathversion{normal}$\Delta$}: point estimate. The second column prints the
  seed-fixed reading (for a pool, the AUC of the per-video seed mean, a
  three-model ensemble); the last column prints the
  difference in mean single-model AUC, with each arm's
  seeds resampled and then the 140 source-identity clusters (an s42 arm is one
  run, so rows~1--4 resample the clusters only). Weights per arm:
  Table~\ref{tab:config}. The last row was not
  adjudicated and lies outside every multiplicity family.}
  \label{tab:contrasts}
  \footnotesize
  \setlength{\tabcolsep}{2pt}
  \begin{tabular}{@{}R{0.208}rlR{0.075}R{0.18}R{0.202}@{}}
    \toprule
    contrast @ c40 & $\Delta$ & 95\% CI & rule & verdict & post hoc: seeds${+}$source clusters resampled, $\Delta$ [95\% CI] \\
    \midrule
    SpatialBase $-$ CAFRL as specified (s42) & $+0.0366$ & \ci{+0.020}{+0.054} & registered & significant --- the gate fires & $+0.0366$~\ci{+0.0182}{+0.0569} \\
    Fair-N $-$ CAFRL$-$adv (s42) & $+0.0396$ & \ci{+0.0231}{+0.0570} & pre-\zb specified & significant --- the fair recipe recovers the deficit & $+0.0396$~\ci{+0.0222}{+0.0580} \\
    Fair-N $-$ SpatialBase (s42) & $-0.0004$ & \ci{-0.0092}{+0.0080} & pre-\zb specified & no detectable difference & $-0.0004$~\ci{-0.0095}{+0.0080} \\
    Fair-W $-$ SpatialBase (s42) & $+0.0055$ & \ci{-0.0022}{+0.0134} & pre-\zb specified & no detectable difference (lean, $P{=}0.93$) & $+0.0055$~\ci{-0.0026}{+0.0137} \\
    \textbf{Fair-N pool $-$ SpatialBase pool (3 vs.\ 3)} & $\mathbf{-0.0009}$ & \ci{-0.0066}{+0.0045} & pre-\zb specified & \textbf{no detectable difference}; post hoc equivalence reading: \SM~\S\ref{app:stats} & $-0.0002$~\ci{-0.0105}{+0.0112} \\
    \textbf{Fair-W pool $-$ SpatialBase pool (3 vs.\ 3)} & $\mathbf{+0.0022}$ & \ci{-0.0035}{+0.0080} & pre-\zb specified & \textbf{no detectable difference}; post hoc equivalence reading: \SM~\S\ref{app:stats} & $+0.0053$~\ci{-0.0042}{+0.0171} \\
    SpatialBase (s42) $-$ TrunkCtrl pool & $-0.0038$ & \ci{-0.0116}{+0.0039} & pre-\zb specified & no detectable difference between the two spatial controls & $+0.0054$~\ci{-0.0040}{+0.0156} \\
    TrunkCtrl pool $-$ spatial-residual (s42) & $+0.0489$ & \ci{+0.0294}{+0.0705} & pre-\zb specified & significant (TrunkCtrl above the shipped-recipe arm) & $+0.0397$~\ci{+0.0196}{+0.0625} \\
    \midrule
    Fair-N pool $-$ TrunkCtrl pool (3 vs.\ 3) & $+0.0014$ & \ci{-0.0051}{+0.0078} & post hoc & outside the family; CI includes 0 & $+0.0027$~\ci{-0.0080}{+0.0135} \\
    \bottomrule
  \end{tabular}
\end{table*}

Per-arm 3-seed c40 means: SpatialBase $0.9565 \pm 0.0160$, Fair-N $0.9563 \pm 0.0062$, Fair-W
$0.9619 \pm 0.0058$ (95\% $t$-bands, $k{=}3$ seeds). Fair-W's lean
(Section~\ref{sec:protocol}) is never significant and is below this design's
resolution. Its sign depends on the comparator:
each Fair-W seed exceeds SpatialBase's 3-seed mean and eight of the nine
cross-seed pairs are positive, but matched by seed index --- which gives the
two runs the same sampling plan but is not an experimental pairing --- one of
the three is negative ($+0.55$, $-0.12$,
$+1.17$\,pt).

An interval containing zero does not by itself establish equivalence. The
$\pm1$\,AUC-point equivalence margin was chosen post hoc, and any equivalence
reading is conditional on these three trained models per arm and does not
extend to the training procedure; the main text therefore says only that no
reliable incremental benefit was detected. The per-unit equivalence readings,
the margin's justification and both readings of the multiplicity adjustment
are in \SM~\S\ref{app:stats}.

An exploratory, post hoc operating-point reading (\SM~Table~\ref{tab:operating};
thresholds set in-sample on each run's own 140 test reals, so the false-positive
rate (FPR) moves in steps of $1/140 = 0.714$\,pt) gives c40 true-positive rates (95\% intervals) at
5\% FPR of 81.4\% \ci{74.6}{87.9} for SpatialBase, 79.9\% \ci{72.1}{88.3} for
Fair-N and 85.2\% \ci{75.8}{90.6} for Fair-W (means over three seeds), and
68.6\% \ci{55.2}{78.6} for the one run of CAFRL as specified (source identities
resampled only); no contrast or
test was computed on them.

The fair frequency variants are selected and scored on EMA weights and
SpatialBase on raw weights. Under one selection rule for all three arms --- the
baselines' rule on raw weights, applied to Fair-N and Fair-W in an
exploratory analysis added in revision (\SM~\S\ref{app:exploratory};
mean single-model AUC, seeds then source-identity clusters resampled) ---
Fair-N $-$ SpatialBase at c40 was $-0.37$\,pt \ci{-1.57}{0.93} and Fair-W
$-$ SpatialBase $-0.72$\,pt \ci{-2.14}{0.69}: both intervals include zero, as
under the published selection, and Fair-W's point estimate changes sign. The
shipped-recipe arms were re-scored on raw weights as points only, with no
interval, so the recipe contrasts have no interval-based reading under that rule.

\subsection{Generalization checks: neither resolved}\label{sec:ood}

\textbf{Out of distribution} (train FF++ multi-CRF, test elsewhere), read at
three seeds as the mean $\pm$ its 95\% $t$-band: CAFRL as specified is close to its matched
backbone on Celeb-DF~v2 ($0.705 \pm 0.051$ vs.\ $0.686 \pm 0.035$) --- the one
genuinely cross-dataset corpus of the original analyses --- and on
\emph{near-domain} DFD (c23 $0.819 \pm 0.040$ vs.\ $0.825 \pm 0.054$; c40
$0.635 \pm 0.035$ vs.\ $0.650 \pm 0.031$). No paired contrast is computable on
any of these stored cells (an exploratory re-scoring added in revision gives
paired contrasts on Celeb-DF~v2, DFDC and WildDeepfake;
\SM~\S\ref{app:exploratory}), so nothing here is adjudicated: we report them
descriptively, and note that the bands overlap everywhere --- which under this
paper's own standard (Section~\ref{sec:protocol}) is evidence of nothing
beyond the exclusion of a very large effect. (How large: \SM~\S\ref{app:ood}.) On Celeb-DF~v2 the
point estimates in fact order the arms opposite to FF++ --- the reversal
Limitation~(i) warns of, on the one cell that tests it, though not detectably
at this resolution. The larger Xception, a single seed-42 run not comparable
to those bands, is ahead on both DFD levels (0.881/0.688) and level with the replicated arms on
Celeb-DF~v2 (0.702). The four replicated arms sit at 0.635--0.835 rather than the
0.96 of the intra-dataset comparisons (0.635--0.881 including the Xception
row). Full tables in \SM~\S\ref{app:ood}; the arms of the revision campaign
on these test sets and on FaceShifter are in \SM~\S\ref{app:campaign}
(secondary; their contrasts are exploratory).

\textbf{Unseen manipulation} (FaceShifter, held out; $n{=}280$ videos --- one
2020 face-swap method on the same FF++ footage, a narrowly defined held-out
manipulation test, not an evaluation of generalization in general): no
advantage is detected for these implementations, at either recipe. Under the
directional gate, frozen before any fair-variant FaceShifter number existed
(a generalization \emph{claim} attaches only if the paired-$\Delta$ 95\% CI
excludes zero in the favorable direction), the shipped-recipe 3-seed check's
$+2.2$\,pt c40 lean never reached significance, and the fair variants'
single-seed leans were sign-inconsistent. With all seeds pooled both fair
variants sit \emph{below} the backbone --- descriptively, a pooled
disadvantage, significant unadjusted for both fair variants and surviving
multiplicity adjustment for one of them, \emph{given these three seeds per
arm} (Fair-N $-0.0325$, CI \ci{-0.0545}{-0.0114}; Fair-W $-0.0317$,
\ci{-0.0575}{-0.0065}), but significant for neither once the seeds are
resampled as well, because between-seed spread dominates on this axis; we
retire it by the same rule as the leans (\SM~\S\S\ref{app:ood}
and~\ref{app:stats}, with the seed-resampled intervals). Across shipped and
fair recipes, three seeds each, \textbf{no unseen-manipulation advantage was
detected at any recipe tested} under the pre-specified rule; the one
favorable interval that excludes zero, Fair-N's seed-42 lean under the post
hoc primary reading (\SM~\S\ref{app:ood}), replaces no pre-specified verdict
and does not survive Holm adjustment. Neither direction of this axis is
established.

\subsection{The adversarial loss at the fair recipe (confirmatory)}\label{sec:adversary}

\begin{table*}[t]
  \centering
  \caption{The confirmatory family of the revision campaign, registered
  under the tag \texttt{phase6-prereg} before any of its 15 new runs; the 6
  runs of Fair-N and SpatialBase are the published ones (tier: registered, confirmatory). Estimates in AUC points on the FF++ test split; T3 is the
  unitless ratio $R = (\text{E2a} - \text{anchor}) / (\text{SpatialBase} -
  \text{anchor})$ at c40. Primary estimand: the mean over three seeds of each
  seed's single-model video AUC. The 95\% percentile interval and the two-sided
  $p$ come from one hierarchical bootstrap ($B = 10{,}000$; each arm's seeds,
  then the 140 source-identity clusters), and Holm's procedure runs over the
  five tests at family $\alpha = 0.05$; the smallest attainable $p$ is
  $1/(B+1)$, except in the $t$-based last column. {\mathversion{normal}$\Delta$}: point estimate. MDE: the minimum detectable effect of
  the test as specified, 80\% power, at $\alpha = 0.01$ / $0.05$ (T3: the loss
  of the gain, $1 - R$, at the registered grid gap of 9.52\,pt, the closest
  below the observed 9.75\,pt). Sensitivity: the calibrated $t$-test across
  seeds, Holm-adjusted over its own five $p$-values (secondary, never the
  verdict; for T3 it tests E2a $-$ SpatialBase); it agrees with the primary
  reading on all five tests. Every arm has three seeds, and no run of any
  confirmatory arm was stopped. Weights: EMA for E1 and Fair-N, raw for the
  SpatialBase-family arms (SpatialBase, E2a, the anchor, E3a, E3b). The
  ensemble estimand of every row is in \SM~\S\ref{app:campaign}.}
  \label{tab:confirmatory}
  \footnotesize
  \setlength{\tabcolsep}{3pt}
  \begin{tabular}{@{}lR{0.25}rlllR{0.12}ll@{}}
    \toprule
    test & contrast (endpoint) & $\Delta$ or $R$ & 95\% interval & $p$ & Holm $p$ &
      reading & MDE & sensitivity Holm $p$ \\
    \midrule
    T1 & E1 $-$ Fair-N (c40) & 0.87 & \ci{0.02}{1.88} & 0.042 & 0.084 &
      no detectable difference & 1.78 / 1.43 & 0.070 \\
    T2 & E1 $-$ Fair-N (\auccomp) & 0.73 & \ci{0.12}{1.45} & 0.020 & 0.061 &
      no detectable difference & 1.37 / 1.10 & 0.095 \\
    T3 & retention $R$ against $R = 1$ (c40) & 0.88 & \ci{0.75}{1.03} &
      0.10 & 0.10 & no detectable difference & 0.22 / 0.18 & 0.095 \\
    T4 & codec: real H.264 $-$ PSNR-matched JPEG, averaged over c40 exposure
      (c40) & 7.48 & \ci{5.12}{9.94} & 0.00010 & 0.00050 & helps &
      1.46 / 1.18 & $5.6\times10^{-7}$ \\
    T5 & c40-level data in training: with $-$ without, averaged over codec
      (c40) & 1.88 & \ci{0.79}{2.98} & 0.00050 & 0.0020 & helps &
      1.46 / 1.18 & 0.0016 \\
    \midrule
    \multicolumn{9}{@{}R{0.88}@{}}{\emph{Reported with T3 whatever its outcome:} its
      denominator, SpatialBase $-$ anchor at c40, 9.75\,pt \ci{7.19}{12.40}; its
      numerator, E2a $-$ anchor, 8.63\,pt \ci{6.16}{11.15}.} \\
    \bottomrule
  \end{tabular}
\end{table*}

The revision campaign ran five registered tests, fixed before any of their
new runs existed (Table~\ref{tab:confirmatory}); they are confirmatory. Each
result below is the registered sentence for the reading the analysis
returned, with its numbers and plain names filled in and, for T4 alone, its
method named and its interpretation narrowed in Revision~2, and every one of them ends with the registered sensitivity test
(its printed values are Holm-adjusted). T1 and T2 compare E1
(Section~\ref{sec:campaign}) with Fair-N, three seeds each, both on EMA
weights:

Under the fair recipe, adding the adversarial loss produced no detectable
difference in mean single-model video AUC at c40 ($\Delta = 0.87$\,pt,
\ci{0.02}{1.88}, Holm $p = 0.084$). With 3 seeds per arm this test detects
about 1.78\,pt with 80\,\% power at $\alpha = 0.01$ (1.43\,pt at
$\alpha = 0.05$); smaller effects in either direction are not excluded
(calibrated sensitivity test: $p = 0.070$).

Under the fair recipe, adding the adversarial loss produced no detectable
difference in the mean AUC over c28--c40 (\auccomp) ($\Delta = 0.73$\,pt,
\ci{0.12}{1.45}, Holm $p = 0.061$). With 3 seeds per arm this test detects
about 1.37\,pt with 80\,\% power at $\alpha = 0.01$ (1.10\,pt at
$\alpha = 0.05$); smaller effects in either direction are not excluded
(calibrated sensitivity test: $p = 0.095$).

The registration also fixes a health read-out of the adversary's run:
the $\pm1$-level accuracy, the within-$\pm1$-level validation accuracy of the
compression-level estimator, read at the selected epoch of each seed (seeds
42, 1 and 2 in that order; \SM~\S\ref{app:campaign}). With the adversary,
the $\pm1$-level accuracy at the selected epoch was 0.9531 / 1.0000 / 1.0000 (Fair-N: 1.0000 /
1.0000 / 1.0000). Over its 30 epochs, E1's seed-2 run reached a lowest
accuracy of 0.3672 (epoch~2), below the registered reference of 0.7; the
lowest values of seeds 42 and 1 were 0.8359 and 0.9805, and at no selected
epoch was the accuracy below 0.7.

Two exploratory analyses added in revision sit beside T1
(\SM~\S\ref{app:exploratory}; not registered, no $p$-value, no verdict). They
bear on how firmly its reading can be held and change no registered result.
A selection sensitivity: with E1 and Fair-N re-selected by the baselines'
rule on their raw rather than their EMA weights, E1 $-$ Fair-N at c40 was
0.06\,pt (\ci{-1.13}{1.28}; raw weights), and the re-selection moved E1
itself by $-1.15$\,pt (\ci{-2.21}{-0.21}; raw $-$ EMA). A lean: on our own
CRF-40 re-encode of the c0 test masters, E1 $-$ Fair-N was 1.51\,pt
(\ci{0.25}{2.92}; EMA weights).

T1, T2 and the read-out reflect one adversarial configuration (one
discriminator, one GRL ramp to $\lambda_{\max} = 0.5$, one loss weight);
others were not tested, and the read-out concerns this implementation only.

\section{Training data and robustness to H.264 re-encoding}\label{sec:data}

Two results carry the training-data story --- read
throughout as robustness \emph{under single-pass H.264 CRF re-encoding}, the
only compression this study applies apart from an exploratory test-time check
with H.265 and AV1 (\SM~\S\ref{app:exploratory}):

\begin{enumerate}
\item \textbf{Multi-level training gave a flatter curve for every
architecture trained both ways.} On FF++-derived single-pass H.264 test
data, all fairly trained multi-CRF models of \SM~Table~\ref{tab:sweep} ---
with or without the tested frequency configurations ---
share the same flat degradation (c23-anchored drop 0.015--0.019 against 0.026
for CAFRL as specified; per arm, with the runs used and the post hoc
c0-anchored drop: \SM~\S\ref{app:sweep}). The c23-trained
literature protocol drops further at c40 (Xception 0.866, F3-Net 0.860, FreqNet
0.768 video-AUC; our implementations of these baselines,
\SM~\S\ref{app:baselines}). Each of the four architectures trained both on
c23 alone and on multiple levels --- Xception, F3-Net and FreqNet
(\SM~Table~\ref{tab:appbench}) and EfficientNet-B0 (the anchor against
SpatialBase, \SM~\S\ref{app:campaignarms}) --- had the flatter curve and the
higher c40 AUC when trained on multiple levels. What this demonstrates is precisely \emph{coverage}: every
heavy evaluation level (c28--c40) lies inside the training distribution, and
the only held-out level, c18, is lighter than anything hard. It is not
evidence of generalization to unseen degradation, and we do not present it as
such; the registered test T3 is the leave-out check (Section~\ref{sec:datatests}).
It is also \textbf{in-corpus}. Every arm of \SM~Table~\ref{tab:appcrossdataset} loses 13--19 points
across the same two CRF levels on near-domain DFD
(Section~\ref{sec:ood}) against 2--5 points on FF++, so the c23-anchored drop computed there would read 0.13--0.19 on the AUC scale, not
0.015--0.019. Multi-CRF training bought these arms a flat curve \emph{on the corpus the
diversity was drawn from}; whether it flattens the curve elsewhere is
untested, and these are two-level drops rather than sweeps, no CRF ladder
having been generated outside the FF++ family.
\item \textbf{Real H.264 outperforms JPEG at matched mean PSNR.} The registered test
T4 carries this claim (Section~\ref{sec:datatests}). The original, unmatched
comparison points the same way but is weaker evidence: training Xception with synthetic
JPEG-quality augmentation --- the common shortcut --- reaches only 0.890
\ci{0.861}{0.916} at real H.264 c40 where identical training on real CRF
variants reaches 0.963 \ci{0.948}{0.976}, $+7.3$\,pt with non-overlapping CIs
(both flip-only, \textbf{one run each}; post hoc, outside the confirmatory
family of Section~\ref{sec:protocol}; the one contrast read on marginal
rather than paired intervals, as Section~\ref{sec:protocol} flags; its
decision rule and a conservative unpaired test: \SM~\S\S\ref{app:baselines}
and~\ref{app:stats}). In both designs the JPEG-trained arms scored lower at
heavy compression; tested on real H.264, they
carry the distribution-shift result behind claim~C3 (Limitation~(iv)).
\end{enumerate}

A note the adversarial-invariance line should weigh: on multi-CRF data, in this design, the
compression-discriminative features the GRL is designed to remove are
\emph{useful}, and in this implementation, at the shipped recipe, the adversary added nothing
detectable while silently degrading its own conditioning signal (D6,
Section~\ref{sec:audit}); whether an adversary works against that mechanism in
other configurations or in general is untested (Section~\ref{sec:adversary}).

\subsection{Which part of the training data matters (confirmatory)}\label{sec:datatests}

T3--T5 (Table~\ref{tab:confirmatory}) ask which part of SpatialBase's
multi-level training data its c40 result depends on (arms:
Section~\ref{sec:campaign}; three seeds each, raw weights). In the registered
wording:

Trained without c40, SpatialBase kept $R = 0.88$ (\ci{0.75}{1.03}) of the
c40 gain; $R$ is not detectably below 1 (Holm $p = 0.10$). At the observed gap
of 9.75\,pt this test detects a loss of about 0.22 of the gain with 80\,\%
power at $\alpha = 0.01$ (Table~\ref{tab:confirmatory}); a smaller loss is not excluded
(calibrated sensitivity test: $p = 0.095$).

At matched mean crop PSNR, training on real H.264 levels gave a higher mean
single-model c40 AUC than training on JPEG versions of the same c0 frames, by
7.48\,pt averaged over c40 exposure (\ci{5.12}{9.94}, Holm $p = 0.00050$). In
this design (EfficientNet-B0, FF++, JPEG matched on mean PSNR), real H.264
outperforms JPEG at matched mean PSNR (calibrated sensitivity
test, a Holm-adjusted analytic $t$-test: $p = 5.6\times10^{-7}$).

This compares two specified procedures for generating training data, not
codec identity as an independent cause: matching mean PSNR does not equalize
the distortions, and the arms' crops differ in geometry (mean landmark shift
1.35\,px at c40; effect on AUC not measured). On a train-split sample
(6{,}400 crops per level; \SM~Table~\ref{tab:codec-descriptors}), per-crop
PSNR SD was larger for H.264 at every level; at c40, 30.61 $\pm$ 0.79\,dB for
JPEG (QF 10) and 30.76 $\pm$ 2.01\,dB for H.264 (mean $\pm$ SD), mean SSIM
0.890 and 0.907 (descriptive); equal mean squared error can hide very different errors \cite{wang2004ssim}, and
reconstruction fidelity and task utility are reported as separate quantities in a study that
measures both \cite{pavat2026vlc}.

Including the c40-level data in training (H.264 c40 or its matched JPEG
quality) raised the mean single-model c40 AUC by 1.88\,pt averaged over codec
(\ci{0.79}{2.98}, Holm $p = 0.0020$). Robustness at c40 depends in part on
exposure to that quality level in training (calibrated sensitivity test:
$p = 0.0016$).

The gap in the T3 sentence and the ratio's numerator are in the last row of
Table~\ref{tab:confirmatory}. Four registered secondary analyses sit beside
these tests and change none of their readings: the symmetric-selection
control (Section~\ref{sec:campaign}), the interaction of codec and c40
exposure (descriptive), E2b (descriptive) and the same contrasts on the
other test sets (exploratory); their registered sentences and the direction
fixed in advance are in \SM~\S\ref{app:campaignsecondary}.

An exploratory leave-one-seed-out analysis added in revision (not
registered; \SM~Table~\ref{tab:loso}) re-estimates T1, T2 and T5 with one seed
dropped from every arm, and changes no registered reading. T1 reads as
registered in all 9 two-seed subsets. T2 reads differently in the same 2 of
9 at each of three bootstrap random seeds, both dropping Fair-N's seed 42. T5 reads
differently in 12 of 81 at one bootstrap seed and in 5 and 10 at two others,
near the Holm line; every T5 interval stays above zero.
An exploratory two-way bootstrap added in revision (not registered;
\SM~\S\ref{app:twoway}) resamples target and source identities
independently. Each test identity has one partner: 140 of 19{,}600
target-by-source cells are occupied. With each real video as a diagonal
cell, the primary rule of this analysis, intervals widen 1.41--1.61 times
against the seed-resampled source-identity unit, and the scheme over-covers on this grid
(99.3--100\% at nominal 95\% in seed-free simulations). T5 then reads no
detectable difference at three bootstrap seeds (Holm $p = 0.113$ over the
five two-way $p$-values; registered 0.0020), though its interval, 1.88\,pt
\ci{0.20}{3.72}, excludes zero; with real videos weighted by target, it
reads as registered (Holm $p = 0.019$). No other registered reading
changes, and T5's registered reading stands. A post hoc check, with seeds
resampled: the four other units of Section~\ref{sec:protocol} give intervals at most 1.054 times as wide
as the source-identity unit on T1, T2, T4, T5 and six fair-recipe rows (listed in \SM~\S\ref{app:twoway}), and
the identity-pair unit, which holds both roles of an identity in one
cluster, changes no reading.

An exploratory analysis added in revision (\SM~\S\ref{app:exploratory}; not
registered, no verdict) bounds where these results hold. On a fixed
1{,}000-video DFDC subset the multi-level SpatialBase scored below the
c23-only anchor: SpatialBase $-$ anchor was $-11.68$\,pt (\ci{-14.98}{-8.57}).
On the WildDeepfake test split the same difference was $-1.92$\,pt
(\ci{-7.13}{2.74}), an interval that includes zero; because the DFDC and
WildDeepfake clusters are approximate, these intervals may be too narrow. On Celeb-DF~v2 the T5
form, the c40 level in training, was $-3.15$\,pt (\ci{-6.54}{-0.04}) on the
original crops and $-2.97$\,pt (\ci{-6.46}{0.16}) on crops re-cut with a
face-track procedure. The robustness results of this section are therefore
stated for FF++-derived test data under single-pass H.264 re-encoding.

\section{Discussion}\label{sec:discussion}

\subsection{Why publish a negative?}
Because the field's default response to
compression is architectural, and our controlled evidence, for this design,
found no detectable gain from the architecture's frequency streams once the
training recipe was matched, while the \emph{apparent} deficit of CAFRL as
specified was removed by changing the recipe package as a whole
(Section~\ref{sec:conclusionfinal}, findings 1--3). Both facts are
cautionary in opposite directions: architecture enthusiasm can be empty, and
negative results can be artifacts. The only way we know to distinguish these
is the discipline used here. The genre itself needs the precedent: negative
results remain structurally under-published in machine learning despite
explicit calls to normalize them \cite{karl2024negative}, and the systemic
cost of the missing negatives is measurable --- in a neighboring applied-ML
field, 79\% of papers claiming to beat standard methods compared against weak
baselines, with reporting biases suppressing the failures
\cite{mcgreivy2024weak}. Our augmentation and capacity confounds
(Section~\ref{sec:confounds}) are precisely that weak-baseline failure mode,
caught in flight only because the controls were matched before the comparison
was believed.

\subsection{What made the negative trustworthy}
Three practices did the work --- registered and pre-specified gates, matched controls, and
auditing our own negative as an adversary would --- and the practice list at
the end of this section states each as something a reader can adopt. What is
worth pricing here is that they were cheap: the STOP-gate saved
${\sim}270$\,h of ablation training on a configuration that had failed its advancement criterion, closing the
researcher-degrees-of-freedom loophole frozen analysis plans exist to close
\cite{simmons2011false,nosek2018preregistration}; the two confounds would each
have flipped the headline; and the audit and its re-tests cost ${\sim}35$\,h
plus ${\sim}31$\,h of confirmation seeds. We propose the audit-the-negative
protocol as a template: a negative about a fashionable architecture should be
published with the audit that tried to kill it. What the original analyses
and the revision campaign cost in compute, and that none of the original
runs failed or was aborted, are recorded in \SM~\S\ref{app:campaign}.

\subsection{Where the next lever probably is}
The main open question is the standalone frequency head: scored alone on the
test split it reaches 85.27 AUC points at c40 at the selected checkpoints
(exploratory), yet the fused
models showed no detectable difference from the spatial control
(Section~\ref{sec:redundancy}). The manuscript supports several explanations
and does not decide between them: the trunk may already carry the head's information;
the fusion (CMAM or the $\gamma$-residual) or the one band-gate configuration
tested may not exploit it; the matched recipe, not tuned for the frequency
side, may not train the fused path to use it (Fair-W selects epochs 5--7 of
30);
and what the head uses is not isolated to frequency (no pixel-domain
control). The input also bounds the reading: the block-DCT and FFT-phase streams see
decoded, warped face crops, not the codec's own coefficients or bitstream, so
the encoder's $8{\times}8$ transform grid is not preserved in what they
analyze (Section~\ref{sec:intro}, Limitation~(ii)). The neutral result
therefore concerns this decoded-domain representation; it does not test
coefficient-domain, grid-aligned frequency features.

One side-observation points past
compression --- and it is the one recommendation in this paper resting on a
number the reader must take on our word, so we grade it before we use it: a
single-seed observation whose paired contrast is recorded in the campaign
journal and is \emph{not} re-derivable from the released artifacts, because
that follow-up did not persist its per-video scores (audit defect D8). With
that grade attached: the highest-capacity spatial model in this study,
Xception, wins every intra-dataset cell yet gives up ${\sim}11$\,pt to the
smaller models on the held-out FaceShifter manipulation at c23: in this one-seed check, capacity spent on the manipulations
it was trained on bought nothing on one it had never seen. The contrast is $\Delta$(CAFRL as specified $-$ Xception-mq) at FaceShifter
c23 $= +0.113$, 95\% CI \ci{+0.064}{+0.162}, $n{=}280$; we print it with the
provenance above rather than silently or not at all. That is consistent, on FF++-derived test data, with
the compression lesson on a second axis --- the reading, untested here, that
what binds is what the training distribution \emph{covers} --- which makes multi-manipulation (and multi-generator)
training the natural successor to multi-CRF training. It also meets the
frequency-generalization literature head-on: spectral features are argued
there to mitigate exactly this texture-overfitting failure
\cite{luo2021generalizing,liu2021spsl,tan2024freqnet}, and our own fair
frequency variants, as implemented and without a hyperparameter search, did
not mitigate it either (Section~\ref{sec:ood}).

\subsection{Where these systems stand}\label{sec:standing}
 A neutral verdict is only as
interesting as the system it is rendered on, so it is worth saying where ours
sits. SpatialBase is not the strongest detector in this study: Xception-mq
clears it by 0.4\,pt at c40 augmentation-matched (0.9631 vs.\ 0.9590) and by
0.9\,pt with the standard policy, at five times the parameters
(\SM~\S\ref{app:baselines}). Out of distribution our replicated arms land at
0.647--0.705 on Celeb-DF~v2, which is \textbf{below} the published record at
every comparison we can find: the single-harness FF++-trained CNN detectors of
\cite{yan2023deepfakebench} run 0.73--0.765 there (three to twelve points
above our four replicated arms, gaps 2.5 to 11.8 points; frame-level
where ours are video-level), and the cross-dataset
frontier, video-level like ours and printed in percent as in those papers,
is 82.4 for LipForensics
\cite{haliassos2021lipforensics}, 86.9 for RealForensics
\cite{haliassos2022realforensics} and 93.2 for self-blended-image training
\cite{shiohara2022sbi}, all FF++-trained. Those buy their generalization with
mouth-motion pretraining, audio-visual self-supervision and synthetic blending
supervision, none of which is in scope here --- but the gap from our best replicated arm is 12 to 23 AUC
points and should be read as such. The neutral verdict is therefore
rendered \emph{at one operating point}: a 4--21\,M CNN trained on
FaceForensics++ with multi-CRF diversity, ${\approx}0.96$ intra-dataset and
${\approx}0.70$ cross-dataset --- a mid-field system, not a frontier one.
Whether a frequency path earns its keep on a substantially stronger or weaker
system is not something these experiments can say. The intra-dataset yardstick
is calibrated internally (\SM~\S\ref{app:baselines}).

Robustness to familiar compression and generalization are different axes
here. The flat curves (Fig.~\ref{fig:sweep}) are robustness within one
family: FF++-derived test videos re-encoded by single-pass H.264. Outside
that family the replicated arms sit at 0.647--0.705 on
Celeb-DF~v2 (Section~\ref{sec:ood}), and in the exploratory analysis the
multi-level-trained SpatialBase scored $11.68$\,pt below the
c23-only anchor on a DFDC subset, with an interval including zero on
WildDeepfake (Section~\ref{sec:datatests}). The compression results are
therefore not evidence of cross-dataset generalization, and no
compression-stratified test exists outside the FF++ family
(Limitation~(iv)).

\subsection{Reconciling the published positives}
 Several works report that
frequency or compression-invariance machinery helps, and this study reports
that, for these implementations at this recipe, it did not. The two are not in direct contradiction, and the reason is
visible in what those works controlled. \SM~Table~\ref{tab:controls} audits the
six works we cite most directly on two questions: was the baseline they beat
trained on the \emph{same} compression-diverse data, and was it matched in
size? None of the five detection studies among them reports a parameter count or a FLOP
figure, and none trains a plain spatial backbone on compression-diverse data
(the two patterns in detail: \SM~\S\ref{app:controls}). That is not a charge of error: their comparisons answer the question they
posed, whether the mechanism helps \emph{at a fixed compression level}. It
does mean our result and theirs are compatible under a hypothesis we can state
but did not test --- that compression-invariance machinery substitutes for
training-data diversity and earns its keep exactly where that diversity is
missing. Every positive in the table was measured in that regime; every
frequency or invariance arm in Table~\ref{tab:config} was not. What was observed, apart from that hypothesis: \emph{with
multi-level H.264 training data, at the matched recipe, the
tested frequency configurations (Table~\ref{tab:config}), or adding the
adversarial loss at one strength, gave
no detectable difference at the resolution of this experiment --- on
FF++-derived test data (on the exploratory DFDC subset multi-level training
did not help; Section~\ref{sec:datatests}) and in our setting
(FaceForensics++-family, GAN-era, single-pass H.264 data; diffusion content,
transcoding chains and unknown codecs untested), and for these
implementations; for training on a single level, this study says
nothing.} Deciding
between that hypothesis and the alternative --- that the machinery adds
nothing at either data regime --- needs a single-level-plus-invariance arm we
did not run, and we flag it \textbf{untested} rather than argue it.

\subsection{Limitations}
Each item is stated here; its detail is in \SM~\S\ref{app:limits}.
(i) Training and intra-dataset evaluation are
FaceForensics++-family. DFD is \emph{near-domain} rather than cross-dataset,
so the only genuinely cross-dataset check outside the exploratory analyses
is Celeb-DF~v2; DFDC \cite{dolhansky2020dfdc} and WildDeepfake
\cite{zi2020wilddeepfake} enter only an exploratory analysis
(\SM~\S\ref{app:exploratory}). Ties at 0.70 AUC leave far more room for an
ordering to reverse than ties at 0.96 do, and we cannot exclude that a broader
out-of-distribution battery would separate arms that are indistinguishable
here.
(ii) ``Frequency'' in this paper means one instantiation: $8{\times}8$
block-DCT and FFT-phase streams, an estimator-conditioned band gate, and CMAM
or $\gamma$-residual fusion, trained at the matched recipe; other families
were not tested. Our block-DCT is also codec-naive by construction (the
encoder's $8{\times}8$ transform grid is neither aligned with nor preserved
in the crop), and whether what \emph{separates} real from fake survives that
was not analyzed.
(iii) The \emph{fair} recipe is the baseline's recipe applied symmetrically,
which is what makes the comparison fair and also what bounds it: no
hyperparameter search was run on the frequency side, every fair arm of the
original analyses runs with the adversary off, and the adversary was tested
at the fair recipe only by one arm at one strength ($\lambda_{\max} = 0.5$,
three seeds; T1 and T2, Section~\ref{sec:adversary}). Conclusions about the
frequency variants hold for these implementations at this recipe; those
about the reproduced baselines are restricted to those implementations
(\SM~\S\ref{app:baselines}).
(iv) \emph{Codec, and the corpus $\times$ compression cell.} Compression here
is single-pass H.264 CRF re-encoding; H.265 and AV1 enter only an
exploratory test-time analysis, multi-pass H.264 encoding and platform
transcode chains are untested, and no compression-stratified evaluation
exists outside the FF++ family, so C2's neutrality across compression levels
and C3's training-data result are both untested cross-dataset.
(v) \emph{Scale.} The trunk here is one 4.0\,M 2019-era CNN;
foundation-model trunks \cite{ojha2023universal,liu2024fatformer} are
untested in either direction, and our own out-of-distribution table says the
axis is live (\SM~Table~\ref{tab:appcrossdataset}). Whether a bolted-on
frequency branch matters less or more as the trunk grows stronger is the
obvious next experiment, not something this study settles.
(vi) \emph{Era.} Every manipulation evaluated here is GAN- or graphics-era;
no diffusion-generated video is tested, so the neutral verdict is era-bound.
The protocol is not, and the compression lesson compares training-data
procedures, not generators, and is stated for FF++-derived, single-pass H.264
test data.
(vii) The neutral verdicts rest on three seeds per arm, with the resolution
of Section~\ref{sec:protocol}; their post hoc $\pm1$\,pt reading holds only
\emph{given those seeds} (\SM~\S\ref{app:stats}).
Mapped onto the four claims: C2 and C4 rest on FF++ plus one held-out
manipulation, near-domain DFD and cross-dataset Celeb-DF~v2 (with out-of-distribution paired contrasts,
and DFDC and WildDeepfake, only in the exploratory
analyses), at single-pass
H.264 CRF, on GAN- and graphics-era content and non-foundation trunks; C3 adds
the shift between JPEG training and H.264 testing, with H.265 and AV1 only at
test time and only exploratory; C1, the protocol, is
method-level and assumes nothing about data, manipulation family, codec or
era.

\begin{table*}[t]
  \centering
  \caption{The findings of Section~\ref{sec:conclusionfinal}, each at or
  below its wording there, with the contrast and estimand it rests on, the
  factors that differ jointly between the contrast's two sides
  (Table~\ref{tab:config}) and its evidence tier, written ``timing, role''
  (timing: registered, pre-specified or post hoc; role: confirmatory,
  secondary, exploratory or descriptive; defined in
  Section~\ref{sec:tiers}). One word stands alone for \emph{post hoc}
  (outside every family, no role), \emph{exploratory} and \emph{descriptive}
  where the paper prints no timing. Row~$k$:
  Table~\ref{tab:contrasts}; T1--T5: Table~\ref{tab:confirmatory}. The
  numbers are in the sources. C1 (the protocol) and C4 (the benchmark
  sweeps, descriptive) are contributions of method and measurement, not
  findings, so they have no row.}
  \label{tab:claims}
  \scriptsize
  \setlength{\tabcolsep}{2pt}
  \begin{tabular}{@{}R{0.17}R{0.27}R{0.215}R{0.15}R{0.13}@{}}
    \toprule
    finding & contrast and estimand & factors changed together & evidence
      tier & source \\
    \midrule
    1.~Configuration result: the original implementation performed worse
      than its spatial comparator, at the shipped recipe (as configured, one
      seed; not an architectural effect) &
      SpatialBase $-$ CAFRL as specified (row~1): $\Delta$ video AUC at c40,
      paired on test videos, one seed-42 run per arm, video bootstrap &
      frequency path, fusion, aux.\ losses, adversary, trunk tap, parameters,
      recipe; weights (raw vs EMA) &
      registered, confirmatory (the STOP-gate) &
      Section~\ref{sec:stopgate}; Tables~\ref{tab:stopgate},
      \ref{tab:contrasts} \\
    \addlinespace
    2.~Recipe sensitivity: the matched training recipe recovered that
      deficit; of the changes reverted singly, only the backbone learning
      rate's c40 interval excluded zero, and this replicated on two new seeds
      (a sensitivity, not a causal decomposition) &
      Fair-N $-$ CAFRL$-$adv (row~2), then Fair-N $-$ SpatialBase (row~3):
      $\Delta$ video AUC at c40, paired on test videos, seed-42 runs, video
      bootstrap. Replication: reverted $-$ Fair-N at c40, arm means over seeds
      1 and 2, seeds then source-identity clusters resampled &
      row~2: the recipe as a whole, including loss weights, curriculum,
      learning rates, freezing, training length and GPU count
      (\SM~\S\ref{app:campaignarms}), one architecture; matched selection. Row~3: as in
      finding~3 &
      pre-specified, confirmatory. One change reverted at a time (E13):
      registered, descriptive, one seed. Backbone-LR replication: exploratory,
      two new seeds, decided by a rule frozen under tag
      \texttt{phase7-prereg} before the runs &
      Sections~\ref{sec:recipeartifact}, \ref{sec:e13};
      Tables~\ref{tab:contrasts}, \ref{tab:e13} \\
    \addlinespace
    3.~Neutral result: the tested frequency configurations, not separately
      tuned and run on the matched, baseline-derived recipe, showed no
      detectable difference at the resolution of this experiment from the
      spatial control, under the evaluated configurations &
      Fair-N, Fair-W $-$ SpatialBase at c40. (a)~Mean single-model AUC,
      three seeds per arm, seeds then source-identity clusters resampled (the
      primary reading); (b)~$\Delta$ paired on test videos, video bootstrap, of the
      seed-pooled scores, conditional on the three seeds run (rows~5, 6), and
      of the seed-42 runs (rows~3, 4) &
      Fair-N: frequency path, CMAM, aux.\ loss, trunk tap, parameters. Fair-W:
      frequency path, $\gamma$-residual fusion, aux.\ losses, parameters. Both: weights and training
      driver (EMA vs raw; Table~\ref{tab:config}) &
      (a)~post hoc; (b)~pre-specified, confirmatory. $\pm1$\,pt reading:
      post hoc, conditional on the seeds run. Standalone head and one-rule selection: exploratory (narrow
      only) &
      Sections~\ref{sec:redundancy}, \ref{sec:seeds};
      Table~\ref{tab:contrasts} \\
    \addlinespace
    4.~One-configuration neutral result: the adversarial loss, at one
      strength, showed no detectable effect at the matched recipe &
      E1 $-$ Fair-N (T1, T2): difference in mean single-model AUC at c40 and on
      \auccomp{}, three seeds per arm, seeds then source-identity clusters
      resampled; Holm over T1--T5 &
      the adversary only; matched selection &
      registered, confirmatory. Raw-weight re-selection: exploratory.
      Shipped-recipe ablation and the D6 read-out: descriptive &
      Sections~\ref{sec:originalchain}, \ref{sec:adversary};
      Table~\ref{tab:confirmatory} \\
    \addlinespace
    5a.~Registered data result: real H.264 outperforms JPEG at matched mean
      PSNR (T4) &
      E3a, E3b against SpatialBase, E2a: difference in mean single-model c40
      AUC averaged over c40 exposure; bootstrap as in T1 &
      how the training levels were generated (real H.264 vs JPEG at matched
      mean crop PSNR); matched within each pair &
      registered, confirmatory. Original JPEG-augmentation comparison: post
      hoc, not relied on &
      Sections~\ref{sec:data}, \ref{sec:datatests};
      Table~\ref{tab:confirmatory} \\
    \addlinespace
    5b.~Including c40-level data in training raised mean c40 AUC (T5) &
      with $-$ without c40-level data (SpatialBase, E3a against E2a, E3b),
      averaged over codec; as T4 &
      c40 in training and the validation level (c40 vs c36) jointly; asym. &
      registered, confirmatory. Symmetric-selection control: registered,
      secondary &
      Section~\ref{sec:datatests}; Table~\ref{tab:confirmatory} \\
    \addlinespace
    5c.~Trained without c40, SpatialBase's retained share of its c40 gain
      was not detectably below 1; a smaller loss is not excluded (T3), so
      c40-level exposure is not shown indispensable; read apart from T5 &
      $R = (\text{E2a} - \text{anchor}) / (\text{SpatialBase} -
      \text{anchor})$ at c40 against $R = 1$; as T4 &
      as T5 &
      registered, confirmatory. Symmetric-selection control: registered,
      secondary &
      Section~\ref{sec:datatests}; Table~\ref{tab:confirmatory} \\
    \bottomrule
  \end{tabular}
\end{table*}

\subsection{Implications for practice} If the apparatus is the transferable
part, it should be stated as something a reader can adopt. Five items, each
with where this paper instantiates it.
\textbf{(1) Match the control on data, capacity, augmentation \emph{and
optimization recipe} before you believe a win} --- two of the first three were
live confounds here and each would have flipped the headline on its own
(Section~\ref{sec:confounds}, \SM~\S\ref{app:baselines}), and matching the fourth
mattered too: changing the recipe removed most of
the gap we first attributed to architecture (Sections~\ref{sec:recipeartifact}
and~\ref{sec:e13}). The reusable control is
TrunkCtrl: train the candidate's own trunk, alone, under the comparator's
recipe, and report where it lands between the two arms before attributing
any gap to architecture (Section~\ref{sec:recipeartifact}).
\textbf{(2) Freeze the decision rule, and freeze a comparison \emph{type}
rather than a checkpoint} --- a rule naming ``the strongest baseline'' survives
the comparator you have not built yet; a rule naming a run does not
(Section~\ref{sec:protocol}).
\textbf{(3) Watch two training-health invariants} --- no auxiliary loss should
exceed the classification loss by more than about an order of magnitude, and
any conditioning signal the model consumes should be checked for accuracy
\emph{during} training, not only at design time; both failures were silent
here and both were audit defects (Section~\ref{sec:audit}, D1 and D6), and both
now ship as warnings in the released code.
\textbf{(4) Probe a branch with a standalone head before concluding it is
useless} --- a branch whose removal costs nothing may still be
individually discriminative (ours: Fig.~\ref{fig:mechanism}b). That rules out ``there is
nothing here'' and leaves several live readings --- among them ``the trunk already has this''
and ``the fusion never let it speak'' --- which have different implications
for what to try next; the exploratory score-fusion and error-overlap
read-outs reported here do not separate them (Section~\ref{sec:redundancy},
\SM~\S\ref{app:exploratory}).
\textbf{(5) Audit your own negative as an adversary would, before publishing
it} --- four of the eight defects we found biased the comparison
\emph{against} the hypothesis under test, whose registered gate had just
fired, and the deficit their joint repair recovered was larger than
the effect under study (Section~\ref{sec:audit}, Appendix~\ref{app:ledger}).

\begin{table*}[!t]
  \centering
  \caption{The D1--D8 audit ledger. ``Repaired/resolved by'' names the
  pre-specified re-test (or documentation fix) that closed each defect.
  \emph{Evidence level}: \emph{identified} (in the original runs'
  artifacts); \emph{effect shown} (the repair changed an original
  arm's result, ``jointly with'' a defect repaired with it); \emph{repair
  seen} (read-outs show it working; no test); \emph{correction tested alone}
  (a run or probe changes only that repair, or one part of it; E13's
  curriculum arm also reverts its loss weights); \emph{no effect
  detected} (that comparison found none); in parentheses, the evidence tier as in
  Table~\ref{tab:claims}.
  Row~$k$: Table~\ref{tab:contrasts}.}
  \label{tab:appledger}
  \footnotesize
  \setlength{\tabcolsep}{3pt}
  \begin{tabular}{@{}R{0.30}R{0.23}R{0.31}R{0.12}@{}}
    \toprule
    defect & why it mattered & repaired/resolved by & evidence level \\
    \midrule
    \textbf{D1} --- the L1 consistency loss ran 40--120$\times$ the
    classification loss at the curriculum's compressed-phase onset, in
    \emph{every} arm (including CAFRL$-$adv) &
    no arm ever trained its frequency features free of invariance pressure;
    the inference ``not the GRL $\Rightarrow$ frequency itself'' was void &
    Fair-N (zero invariance losses, flat curriculum). \textbf{A principal
    candidate cause, repaired together with D3, so its share cannot be
    isolated}: $+3.96$\,pt recovered at c40 (Fair-N vs.\ CAFRL$-$adv, seed
    42), significant &
    effect shown, jointly with D3 (row~2; pre-specified, confirmatory) \\
    \addlinespace
    \textbf{D2} --- $\gamma\rightarrow0$ in the residual probe was a
    curriculum-shock collapse locked in by a structural deadlock: at
    $\gamma{=}0$ the frequency path receives zero label gradient &
    the probe could not detect a useful frequency signal even if present;
    ``the model freely discards frequency'' was unfounded &
    Fair-W (auxiliary BCE gives the frequency path label gradient regardless of
    $\gamma$). \textbf{Deadlock broken}: $\gamma$ grows to 0.146, decaying to
    0.031 by epoch 30 rather than collapsing &
    identified; repair seen (descriptive) \\
    \addlinespace
    \textbf{D3} --- optimization never matched: baseline full fine-tune at
    $2{\times}10^{-4}$ from epoch 1 vs.\ CAFRL's 20$\times$ lower backbone learning rate,
    10-epoch freeze, heavy data deferred to the learning-rate tail &
    part of the deficit was recipe, not architecture &
    TrunkCtrl $+$ Fair-N. The fair recipe, with the frequency stream, aux.\ losses and adversary removed, closed most of
    the 0.0451 trunk gap (TrunkCtrl 0.0397 above the shipped-recipe arm, the
    only difference whose interval excludes zero, and 0.0054 below
    SpatialBase; Section~\ref{sec:recipeartifact}), without identifying which
    modification did so (reverted one at a time at one seed, only the backbone
    learning rate gave a c40 interval that excludes zero, and this replicated
    on two new seeds; Section~\ref{sec:e13}); the fair recipe recovers the model-level deficit &
    effect shown, jointly with D1 (row~2; row~8 also differs in frequency path, aux.\ losses, adversary, parameters and weights; pre-specified, confirmatory);
    correction tested alone, one part at a time (E13; registered,
    descriptive); backbone learning rate replicated on two new seeds
    (exploratory) \\
    \addlinespace
    \textbf{D4} --- CAFRL's spatial path taps an intermediate 320-channel map
    (256-d projection), discarding the backbone's final 1280-d representation &
    every CAFRL-with-frequency variant carried a narrower spatial trunk, a possible cost untested until TrunkCtrl &
    TrunkCtrl $+$ Fair-W. \textbf{Closed --- no detectable difference between
    the two implemented spatial controls}: SpatialBase $-$ TrunkCtrl is
    $-0.0038$, CI \ci{-0.0116}{+0.0039}, and $+0.0054$ \ci{-0.0040}{+0.0156}
    under the post hoc primary reading (Table~\ref{tab:contrasts}); they
    differ in more than width (4.0\,M vs.\ 3.68\,M), so this does not isolate
    the width; the frequency path on the full trunk
    still makes no detectable difference &
    no effect detected, with parameters (row~7; pre-specified,
    confirmatory) \\
    \addlinespace
    \textbf{D5} --- split-BatchNorm evaluation forces the clean branch at
    eval (train/eval routing mismatch) &
    could have distorted compressed-level evaluation &
    eval-only routing probe. \textbf{Retired --- immaterial}: oracle-routing
    c40 $\Delta \le 0$ on both probed arms (the clean eval mildly
    \emph{flattered} CAFRL) &
    correction tested alone; no effect detected (pre-specified,
    descriptive) \\
    \addlinespace
    \textbf{D6} --- the GRL silently degraded its own conditioning estimator:
    $\pm$1-bin accuracy $0.88 \rightarrow {\approx}0.4$ under the adversary
    (healthy ${\approx}1.0$ without it) &
    the ``compression-aware conditioning'' mechanism was self-defeating as
    configured (``condition-then-remove'') &
    four-way control pattern across arms; estimator-health warning
    (\mbox{$\pm$1-bin} $<0.7$) added to the training diagnostics &
    identified (read-outs across arms; descriptive) \\
    \addlinespace
    \textbf{D7} --- ``discriminator at chance = invariance working'' was a
    misread: the discriminator had collapsed to a constant predictor &
    invariance-vs-dead-adversary is indeterminate from that run &
    documentation corrected; diagnostic read-out fixed &
    identified (descriptive) \\
    \addlinespace
    \textbf{D8} --- artifact hygiene: four cells lacked per-video prediction
    files; a seed-band constant hardcoded for $k{=}3$; an argument-order bug
    in the run aggregator &
    reproducibility of the reported cells &
    per-video predictions backfilled from frozen checkpoints (bit-exact);
    $t$-quantile generalized ($k{=}3$ unchanged); aggregator fixed with
    tests; all evaluation drivers committed &
    identified (descriptive) \\
    \bottomrule
  \end{tabular}
\end{table*}

\section{Conclusion}\label{sec:conclusionfinal}

We asked whether the specific CAFRL frequency pathways provide incremental
discrimination over appropriately controlled spatial architectures when
training includes multiple H.264 compression levels
(Section~\ref{sec:intro}). For this design, whose frequency-stream variants
were not separately tuned and ran on a baseline-derived recipe, we detected
none on FaceForensics++ test data under the evaluated configurations --- no
detectable difference at the resolution of this experiment. The five findings
below rest on different evidence and carry different limits, and we keep them
apart; each names its evidence type, and Table~\ref{tab:claims} gives each
one's estimand, jointly changed factors and evidence tier. CAFRL-family numbers use EMA
weights, the SpatialBase family raw weights.

\begin{enumerate}
\item \textbf{Configuration result: the original implementation performed
worse than its spatial comparator.} Under the registered
STOP-gate, SpatialBase $-$ CAFRL as specified at c40 was $+0.0366$, 95\% CI
\ci{+0.020}{+0.054}, paired on test videos (Section~\ref{sec:stopgate}) --- as
configured, one seed, not an architectural effect. \emph{Limit:} one run per arm at seed 42, as
the gate adjudicated it (the two further shipped-CAFRL seeds lie within 0.002
of that run at c40), and at the shipped recipe only.
\item \textbf{Recipe sensitivity: the matched training recipe recovered
that deficit.} With
CAFRL's own components and the adversary off, the fair recipe gained
$+0.0396$, CI \ci{+0.0231}{+0.0570}, over the shipped-recipe arm CAFRL$-$adv,
and Fair-N then showed no detectable difference from SpatialBase ($-0.0004$,
CI \ci{-0.0092}{+0.0080}); both seed 42, post-audit pre-specified
(Section~\ref{sec:recipeartifact}). \emph{Limit:} the recipe was changed as a
package. With one change reverted at a time in a one-seed descriptive ablation
(registered, E13), only the backbone learning rate gave a c40 interval that
excludes zero (Section~\ref{sec:e13}). That result replicated on two new seeds
under a rule frozen before those runs (exploratory; no $p$-value): reverting
the backbone learning rate changed c40 AUC by $-3.06$\,pt against Fair-N
(\ci{-4.85}{-1.42}; reverted minus Fair-N, arm means over seeds 1 and 2, seeds
and source-identity clusters resampled), and by $-3.39$\,pt
(\ci{-5.12}{-1.78}) over all three seeds, beside it and not decisive. It remains a sensitivity, not a
causal decomposition: the other changes were not replicated, and the changes
were not tested in combination. The recovery shows that the deficit does not establish an architectural failure;
it does not show that every component is irrelevant.
\item \textbf{Neutral result: the tested frequency configurations, not
separately tuned and run on the matched, baseline-derived recipe, showed no
detectable difference at the resolution of this experiment from the spatial
control, under the evaluated configurations.} On the mean
single-model AUC at c40, with seeds and then source-identity clusters
resampled (the post hoc primary reading), Fair-N $-$ SpatialBase was
$-0.0002$ \ci{-0.0105}{+0.0112} and Fair-W $-$ SpatialBase $+0.0053$
\ci{-0.0042}{+0.0171}; the pre-specified contrasts, conditional on the three
seeds run, also include zero (Table~\ref{tab:contrasts}). \emph{Limit:}
these implementations only --- two fusion mechanisms at two spatial-feature
widths of one 4.0\,M trunk. Across seeds the design detects about
1.5--1.7\,pt, so a smaller gain, including one the size of the fairest
published frequency ablation (${+}1.4$\,pt, \SM~Table~\ref{tab:controls}), is not
excluded. The main open question is the standalone frequency head: on the
test split it scores 85.27 AUC points at c40 at the selected checkpoints and
89.99 at epoch 30 (exploratory; Section~\ref{sec:redundancy}), so the
frequency sub-network carries discriminative information, yet the fused models showed no
detectable difference from the spatial control. The head's information may already be in the trunk, or the fusion
may not exploit it, and no pixel-domain control isolates what the head
uses; this study does not decide between these. The comparison is intra-dataset; on the one
held-out manipulation, FaceShifter, neither direction is established:
pooled, both fair variants sat below SpatialBase, significantly (unadjusted) only
conditional on the three seeds run (Section~\ref{sec:ood}). DFDC and WildDeepfake enter only the exploratory
analyses.
\item \textbf{One-configuration neutral result: the adversarial loss, at one
strength, showed no detectable effect at the matched recipe.} At the shipped recipe it added nothing
detectable (CAFRL$-$adv 0.919 against the full model's 0.922 at c40) and
degraded the estimator it conditions ($\pm$1-bin accuracy
$0.88 \rightarrow {\approx}0.4$; D6). Restored at the fair recipe (E1,
registered tests T1 and T2, three seeds per arm), it made no detectable
difference in detection: $\Delta = 0.87$\,pt at c40 (\ci{0.02}{1.88}, Holm
$p = 0.084$) and $0.73$\,pt on \auccomp{} (\ci{0.12}{1.45}, Holm
$p = 0.061$). Its estimator stayed accurate at the selected epochs ($\pm1$-level accuracy
0.9531 / 1.0000 / 1.0000), although one seed fell below the registered
reference early in training (Section~\ref{sec:adversary}). \emph{Limit:} one
arm at one strength ($\lambda_{\max} = 0.5$); these tests detect about 1.78
and 1.37\,pt at 80\% power and $\alpha = 0.01$, and smaller effects in either direction are
not excluded. The result concerns this implementation, not adversarial
compression invariance in general.
\item \textbf{Registered data result: real H.264 outperforms JPEG at matched
mean PSNR.} Against JPEG versions of the same frames at matched mean crop
PSNR, real H.264 training levels gave a higher mean single-model c40 AUC by 7.48\,pt averaged over c40
exposure (\ci{5.12}{9.94}, Holm $p = 0.00050$; T4), and including the
c40-level data raised it by 1.88\,pt averaged over codec (\ci{0.79}{2.98},
Holm $p = 0.0020$; T5); both registered, paired on test videos, three seeds per arm
(Section~\ref{sec:datatests}). \emph{Limit:} trained without c40, SpatialBase
kept $R = 0.88$ \ci{0.75}{1.03} of its c40 gain (T3), not detectably below 1,
with a smaller loss not excluded, so c40-level exposure is not shown to be
indispensable; T5's $+1.88$\,pt is a separate reading, of including
c40-level data, and neither test shows that such exposure is dispensable or
indispensable. The flat degradation curve is shown within the trained
single-pass H.264 levels and on FaceForensics++-derived test data --- the
multi-level-trained arms drop 13--19 points across the same two levels on near-domain DFD, and on
the exploratory DFDC subset the multi-level SpatialBase scored below the
c23-only anchor (Section~\ref{sec:data}). The video compression studied is
single-pass H.264 throughout, with H.265 and AV1 only at test time and only
exploratory. The submitted single-run comparison with JPEG augmentation is
post hoc and unpaired, and this finding does not rest on it. The practical
reading, to train on real H.264 compression levels rather than JPEG
stand-ins, is limited to FaceForensics++-family, GAN-era, single-pass H.264 data;
diffusion content, transcoding chains and unknown codecs were not tested.
\end{enumerate}

What transfers from this study is less a verdict on frequency features than
the procedure that produced these findings. The negative that survived the
internal audit is smaller than the one that entered it --- the deficit of CAFRL as
specified did not survive a matched recipe, and at that baseline-derived
recipe, with the frequency streams not separately tuned, this frequency design
made no detectable difference at the resolution of this experiment under the
evaluated configurations --- and each part of it is stated at the tier
and with the limit its evidence supports.

\section*{Data availability}

The three video corpora of the main analyses are \textbf{not public downloads}: they are
released by their authors under data-use agreements that a requester signs.
FaceForensics++ \cite{roessler2019faceforensics} (from which the DFD /
DeepFakeDetection set is also distributed, at c23 and c40) and Celeb-DF~v2
\cite{li2020celebdf} are each obtained from their authors through those
agreements. The two corpora of the exploratory analysis added in revision
(\SM~\S\ref{app:exploratory}) were obtained as follows: DFDC
\cite{dolhansky2020dfdc} is distributed by its authors for research use, and
we use a fixed subset of its training release; for WildDeepfake
\cite{zi2020wilddeepfake} we took the official test split from a public copy
on Hugging Face (\texttt{xingjunm/WildDeepfake}). We redistribute no video, no frame and no face crop.

What we do release makes the evaluation crops of the main analyses
reconstructible byte for byte from FaceForensics++, DFD and Celeb-DF~v2; for
the exploratory test sets, DFDC and WildDeepfake among them, it ships
per-video predictions but no crop coordinates (\SM~\S\ref{app:repro}).
The crop pipeline and the exact re-encoding command are in
\SM~\S\ref{app:pipeline}.

\section*{Reproducibility statement}

Every decision rule in this paper was frozen before the comparison it governed
was run, at
the tiers Section~\ref{sec:tiers} defines (tags \texttt{phase5-prereg-a1},
\texttt{phase6-prereg} and \texttt{phase7-prereg}; Sections~\ref{sec:tiers} and~\ref{sec:campaign}); the \emph{post hoc}
analyses are labeled as such where they appear.

\noindent\textbf{Checkpoint selection}, per compared arm, is in Table~\ref{tab:config}
(baselines: \SM~\S\ref{app:repro});
the selection rule, video scores and pairing,
what is and is not released, which values each of three reproducibility
levels rebuilds, and the environment and tolerances are in
\SM~\S\ref{app:repro}.

\textbf{Where it is.} The repository --- code, per-video predictions, the
frozen rule texts, the evaluation and analysis harness, and the claims map ---
is \url{https://github.com/Capta1n-n9m0/cafrl-negative}, at tagged release
\texttt{v2.1.0}, the version that matches this manuscript. The archival deposit,
which adds the model checkpoints and the crop coordinates to that tagged tree, is
\url{https://doi.org/10.5281/zenodo.24000169}
(DOI \texttt{10.5281/zenodo.24000169}).

\section*{Generative AI disclosure}

Generative artificial intelligence tools were used in the preparation of this
work, under author direction and with author verification of every resulting
claim, including the internal audit of Section~\ref{sec:audit} --- four
author-directed, tool-assisted internal audit passes, documented in the
\texttt{audit/} directory of the released artifact repository --- and
assistance in code development and analysis tooling. All experiments,
decision rules, analyses and conclusions are the authors' own, and the
authors take full responsibility for the content of this article.

\appendices

\section{The complete audit ledger (D1--D8)}\label{app:ledger}

Table~\ref{tab:appledger} is the full ledger of the internal audit of
Section~\ref{sec:audit}; D1--D4 are the four defects biased \emph{against}
the frequency hypothesis.

\section*{ACKNOWLEDGMENT}
This work was developed within the Center of Data Analytics Research (CeDAR)
and the MegaSec AI Laboratory at ADA University.

\bibliographystyle{IEEEtran}
\bibliography{../refs}

\begin{IEEEbiography}[{\includegraphics[width=1in,height=1.25in,clip,keepaspectratio]{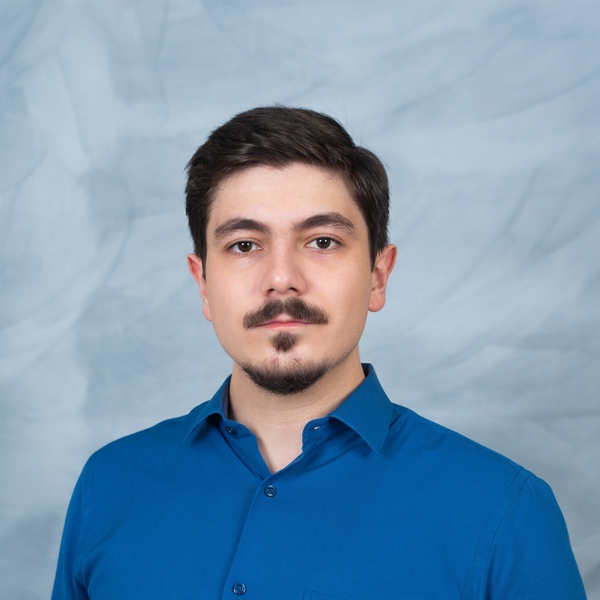}}]{Abbas Aliyev}
received the B.S. degree in computer science jointly from the
French-Azerbaijani University (UFAZ), Baku, Azerbaijan, and the University
of Strasbourg, Strasbourg, France, in 2023, and the M.S. degree in computer
science jointly from ADA University, Baku, Azerbaijan, and The George
Washington University, Washington, DC, USA, in 2025. He is currently
pursuing the Ph.D. degree in system analysis, control, and information
processing with ADA University, where he conducts research at the MegaSec
AI Laboratory within the Center of Data Analytics Research (CeDAR).

From 2022 to 2024, he was a Backend Developer with the French-Azerbaijani
University, and in 2025 he was a Data Scientist with Digital Research Lab,
Baku, where he built signal-processing and machine-learning pipelines for
the energy sector. His research interests include deepfake and
AI-generated-media detection, compression-robust video forensics,
frequency- and temporal-domain representation learning, and rigorous
empirical methodology for machine learning.
\end{IEEEbiography}

\begin{IEEEbiography}[{\includegraphics[width=1in,height=1.25in,clip,keepaspectratio]{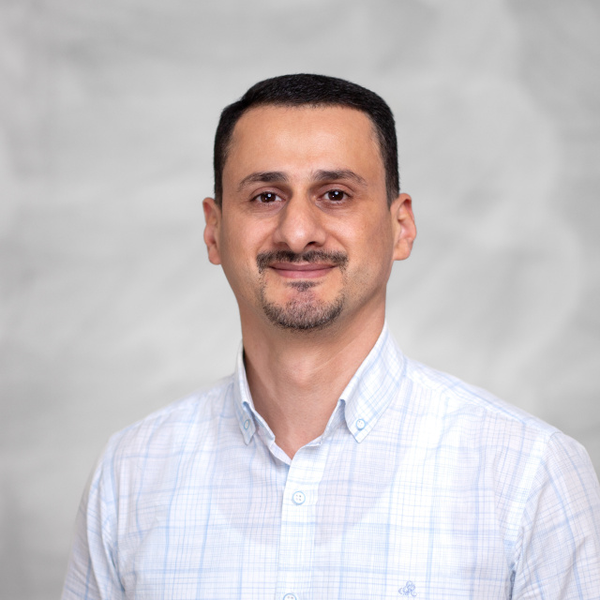}}]{Samir Rustamov}
received the B.S. degree in applied mathematics and cybernetics in 2000 and
the M.S. degree in optimization and optimal control in 2002,
both from Baku State University, Baku, Azerbaijan, and the Ph.D. degree in
mathematical cybernetics from the Institute of Cybernetics, Azerbaijan
National Academy of Sciences (ANAS), Baku, in 2007.

From 2006 to 2016, he was a Researcher with the Institute of Cybernetics,
ANAS. He was a Post-Doctoral Researcher with the School of Electrical and
Computer Engineering, Georgia Institute of Technology, Atlanta, GA, USA,
from 2012 to 2013, and a Visiting Researcher with the Florida Institute of
Technology, Melbourne, FL, USA, in 2015. He is currently an Associate
Professor with the School of Information Technologies and Engineering, ADA
University, Baku, Azerbaijan, where he has been the Program Director of
Computer Science since 2019. He has also been an Adjunct Professor with The
George Washington University, Washington, DC, USA, since 2021, and is the
Head of the AI Laboratory, MegaSec AI Company, Baku. His research interests
include machine learning, natural language and speech processing, and
applied artificial intelligence, with a focus on low-resource languages.

Dr.\ Rustamov received the Best Paper Award at the 8th Machine Intelligence
and Digital Interaction Conference, Poland, in 2020, the Best Teacher Award
from ADA University in 2021, and Best Presentation Awards at the 5th
Artificial Intelligence and Cloud Computing Conference, Japan, in 2022 and
the 7th Natural Language Processing and Artificial Intelligence Conference,
China, in 2026. He serves on the organizing committees of several IEEE and
international conferences on information and communication technologies.
\end{IEEEbiography}

\EOD

\end{document}